\documentclass[10pt,twocolumn,letterpaper]{article}

\usepackage[pagenumbers]{cvpr} %

\usepackage{graphicx}
\usepackage{amsmath}
\usepackage{amssymb}
\usepackage{booktabs}

\usepackage[pagebackref,breaklinks,colorlinks]{hyperref}

\usepackage[capitalize]{cleveref}
\crefname{section}{Sec.}{Secs.}
\Crefname{section}{Section}{Sections}
\Crefname{table}{Table}{Tables}
\crefname{table}{Tab.}{Tabs.}

\usepackage[font=small,labelfont=bf]{caption}
\usepackage{subcaption}
\usepackage{multirow}
\usepackage{graphicx}
\usepackage{epsfig}
\usepackage{tabularx}
\usepackage{mathtools}
\usepackage{tabulary}
\usepackage{booktabs}       %
\usepackage{nicefrac}       %
\usepackage{microtype}      %
\usepackage{booktabs}
\usepackage{pifont}

\usepackage{times}
\usepackage{epsfig}

\makeatletter
\@namedef{ver@everyshi.sty}{}
\makeatother
\usepackage{tikz}

\usepackage[utf8]{inputenc}
\usepackage{graphicx,verbatimbox}
\usepackage{tcolorbox}
\usepackage{color}

\usepackage{tikz}
\usetikzlibrary{shapes.geometric, arrows, arrows.meta}
\usepackage{pgfplots}
\pgfplotsset{width=7.5cm,compat=1.12}
\usepgfplotslibrary{fillbetween}

\newcommand{\cmark}{\ding{51}\xspace}%
\newcommand{\xmark}{\ding{55}\xspace}%

\definecolor{vis}{rgb}{0.2, 0.5, 1.0}
\definecolor{cls}{rgb}{0.8, 0.0, 0.0}
\definecolor{con}{rgb}{0.98, 0.45, 0.5}
\definecolor{lm}{rgb}{0.4, 0.8, 0.4}

\newcommand{\cmmnt}[1]{}

\def\cvprPaperID{8} %
\def\confName{CVPR}
\def\confYear{2023}

\begin{document}

\title{eP-ALM: Efficient Perceptual Augmentation of Language Models} %

\author{
\begin{minipage}{\linewidth}
\begin{center}
 Mustafa Shukor$^1$ \hspace{0.35cm}  Corentin Dancette$^1$  \hspace{0.35cm}
Matthieu Cord$^{1,}$$^{2}$ \\[0.5cm] 
\scalebox{1.}{$^1$Sorbonne University \qquad $^2$Valeo.ai }\\[0.1cm]
\scalebox{1.}{\small{\texttt{\{firstname.lastname\}@sorbonne-universite.fr}}}\\[1cm]
\end{center}
\end{minipage}
}

\maketitle

\begin{abstract}
Large Language Models (LLMs) have so far impressed the world, with unprecedented capabilities that emerge in models at large scales. On the vision side, transformer models (\emph{i.e.}, ViT) are following the same trend, achieving the best performance on challenging benchmarks. With the abundance of such unimodal models, a natural question arises; do we need also to follow this trend to tackle multimodal tasks? In this work, we propose to rather direct effort to efficient adaptations of existing models, and propose to augment Language Models with perception. Existing approaches for adapting pretrained models for vision-language tasks still rely on several key components that hinder their efficiency. In particular, they still train a large number of parameters, rely on large multimodal pretraining, use encoders (\emph{e.g.}, CLIP) trained on huge image-text datasets, and add significant inference overhead. In addition, most of these approaches have focused on Zero-Shot and In Context Learning, with little to no effort on direct finetuning. We investigate the minimal computational effort needed to adapt unimodal models for multimodal tasks and propose a new challenging setup, alongside different approaches, that efficiently adapts unimodal pretrained models. 
We show that by freezing more than 99\% of total parameters, training only one linear projection layer, and prepending only one trainable token, our approach (dubbed eP-ALM) significantly outperforms other baselines on VQA and Captioning across Image, Video, and Audio modalities, following the proposed setup. The code will be available here: \href{https://github.com/mshukor/eP-ALM}{https://github.com/mshukor/eP-ALM}.    

\end{abstract}
\section{Introduction}
\label{sec:intro}

\begin{figure}
    \centering
    \resizebox{\linewidth}{!}{
    \includegraphics[width=\linewidth]{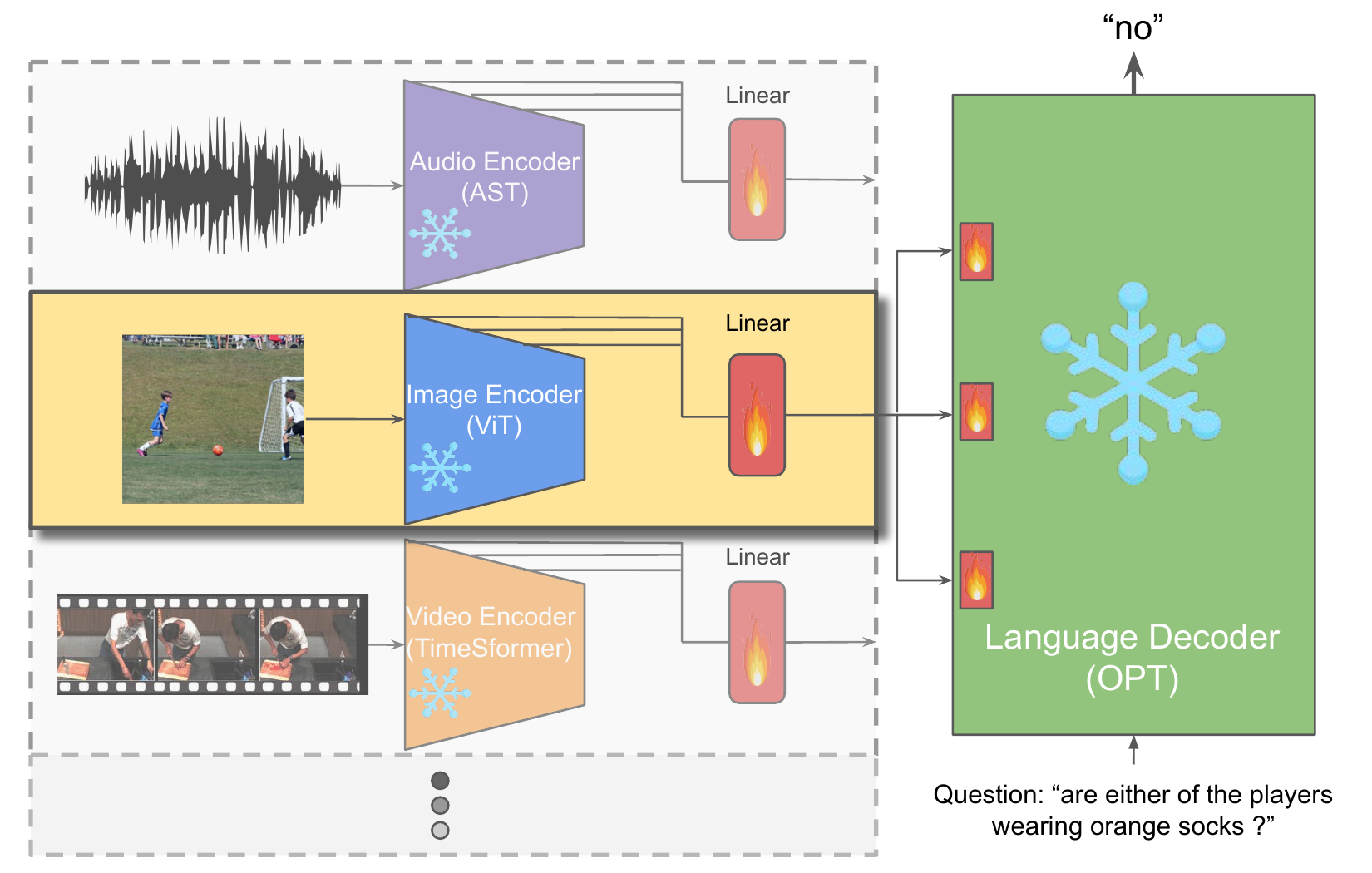}
    }
    \caption{Illustration of eP-ALM to adapt unimodal models for multimodal tasks. The Language Model (Decoder) is augmented with perceptual context to steer its text generation. To condition the decoder on a given modality, the [CLS] tokens are extracted from several layers of a modality-specific encoder and then linearly projected before concatenation at different levels of the language decoder. Only unimodal models are used, and all pretrained modules are kept frozen.
    }
    \label{fig:teaser}
\end{figure}

Going large scale has led to outstanding performances that consistently improve across tasks, modalities, and domains on current benchmarks.  Most of the progress so far has been in the vision and language domains. For Computer Vision, the ViT family \cite{vit} starts from the tiny model with 5M parameters to the enormous ViT-e \cite{chen2022pali} with 4B parameters and the largest ViT-22B with 22B parameters \cite{dehghani2023scalingvit22b}. More captivating, are the scales of Large Language Models (LLMs), such as the BLOOM \cite{scao2022bloom} and OPT \cite{zhang2022opt} families, ranging from hundreds of millions of parameters to 175B, in addition to other models that go beyond 100B \cite{brown2020languagegpt3, smith2022usingmegatron, chowdhery2022palm} up to 1T parameters \cite{fedus2022switch}. These huge scales come with a need for very large pretraining datasets and long training times.

The current prevalent paradigm to solve multimodal tasks, in particular, Vision-Language tasks is to leverage pretrained models, and then further train end-to-end \cite{li2021alignalbef, vicha, singh2022flava, wang2022imagebeit3, chen2022pali} on large image-text datasets. However, the training cost is huge and unaffordable for a large portion of the community, as these approaches still train all model parameters, even after initialization, on a huge amount of data.

With the abundance of  unimodal models, a natural question arises; 

\textit{Do we need also to follow this trend to tackle multimodal tasks? or rather direct effort to efficient adaptations of existing models?}   

Drawing inspiration from the recent work in Augmented Language Models (ALMs) \cite{mialon2023augmentedalm}, in this paper, we advocate for adapting pretrained LMs to solve multimodal tasks. Specifically, by augmenting LMs with perceptual encoders.

Several approaches have deviated from the end-to-end-training paradigm by freezing some pretrained modules and training only the adaptation parameters, such as, additional cross-attention \cite{alayrac2022flamingo}, vision encoder \cite{tsimpoukelli2021multimodalfrozen} and Adapters \cite{eichenberg2021magma}.

Even though these approaches have taken a big step towards more parameter-efficient models, there are still many costly components that hinder their adoption by the large community, such as the training and inference memory and time cost.

In this work we argue that current approaches are far from optimal and it is possible to find more efficient approaches, in terms of the number of trainable parameters, training data, and compute, to adapt pretrained unimodal models for multimodal tasks. A better alignment of visual and language representations might help to devise extremely efficient adaptation approaches. 

To investigate this hypothesis, we go a step further to efficiently leverage LLMs, and propose (1) a new technique to adapt unimodal models by freezing more than 99 \% (up to 99.94\%) of their parameters, alongside (2) a minimal and challenging setup to adapt pretrained unimodal models for Image/Video/Audio-Language tasks (\emph{e.g.}, VQA \cite{goyal2017makingvqav2, msvd_msrvtt}, Image and Audio Captioning \cite{chen2015microsoftcaption, audiocaps}).
In this setup, we favor unimodal-only models, avoiding multimodal pretraining or massively trained multimodal encoders, and considering the typical LLMs architecture as the backbone. All that while freezing as much as possible of model parameters. The approach is illustrated in Fig.\ref{fig:teaser}.

Specifically, we adopt the publicly released OPT model \cite{zhang2022opt} and unimodal encoders (\emph{e.g.}, ViT, TimeSformer \cite{bertasius2021spaceTimeSformer}, AST \cite{gong21b_interspeech_ast}), which are kept frozen. We finetune directly the adaptation parameters on publicly available benchmarks of downstream tasks such as for VQA, GQA, Image Captioning, Video QA, Video Captioning, and Audio Captioning.

Based on this setup we investigate different design choices and propose very efficient approaches backed by the following interesting findings:

\begin{itemize}
\setlength\itemsep{0.5pt}

    \item Training a single linear layer directly on downstream multimodal datasets, and following the same setup, outperforms other work on Image/Video/Audio-Language tasks. With a few additional trainable parameters and a single learned prepended token, we can significantly improve the performance, while respecting a budget of 1\%  of trainable parameters, and keeping almost the same inference cost.

    \item Our approach enjoys better generalization (OOD, Zero-Shot) and is data-efficient (training on 1\% of the data achieves 80\% of performances) with better few-shot results than other approaches.

    \item While reaching good performance with small to mid-scale language models (\emph{i.e}, 350M-2.7B) the improvement still increases by jointly scaling both vision and language models. When scaling both models, we can still outperform other approaches with only 0.06\% of trainable parameters.

    \item Existing approaches do not behave well on the proposed challenging setup, without large multi-modal pretraining. 
    
\end{itemize}

\section{Related Work}
\label{sec:related}

\paragraph{Vision-Language Models (VLMs).} Previously, vision-language tasks have been solved with models heavily customized for the particular task at hand \cite{ke2019reflective, fang2015captions, cadene2019murel, jiang2020defense, hu2019language}. The success in Self Supervised Learning \cite{wu2018unsupervised, grill2020bootstrap, he2020momentummoco, tian2020makes, caron2021emergingdino} and the importance of good initialization have pushed researchers to transfer these ideas to VLMs and started Vision-Language Pretraining (VLP) on large scale video-text \cite{fu2021violet, li2022lavender, wang2022omnivl}, image-text datasets in general domains \cite{lu2019vilbert, chen2020uniter, vicha, li2021alignalbef, kim2021vilt, li2022blip}, as well as specific domains, such as Cooking \cite{shukor2022structuredvlp}, Medical Images \cite{moon2022multi} and Event Extraction \cite{li2022clipevent}. VLP is a step to move away from the burden of customization by having one pretrained model, exploited for several downstream tasks. Recently, we have witnessed impressive work that go a step further towards more unification, by unifying the model, the training objective, and input-output format \cite{wang2022unifyingofa, wang2022imagebeit3, chen2022pali, lu2022unifiedio}. All these models train most of the model parameters, even after initialization, which becomes more and more costly with the current trend in scaling data, model size, and compute \cite{yu2022coca, chen2022pali}. Another approach for VLM is to exploit existing pretrained models by keeping them frozen and training only the adaptation parameters \cite{alayrac2022flamingo, eichenberg2021magma, liang2022modularpromptfuse}. This work advocates for the latter favoring training efficiency in terms of memory and time.

\paragraph{Adapting Language Models.} Large Language Models (LLMs) \cite{brown2020languagegpt3, smith2022usingmegatron, hoffmann2022trainingchinchilla, chowdhery2022palm, zhang2022opt, scao2022bloom} have impressed the world in this last few years, showing unprecedented performance on a myriad of NLP tasks. Scaling LLMs to hundreds of billions of parameters has been motivated by the capabilities that surprisingly emerge \cite{wei2022emergent} at this scale and lead to sudden jumps of relevant metrics on hard downstream tasks \cite{srivastava2022beyondbigbench, patel2022mapping, hendrycks2021measuring}. This generalization ability pushed researchers to start adapting these models for other modalities \cite{tsimpoukelli2021multimodalfrozen, alayrac2022flamingo}, tasks \cite{zeng2022socratic, tiong2022plug, yang2022empirical, hao2022language} and domains \cite{singh2022progprompt}. Currently, most of the focus is concentrated on exploiting LLMs for vision-language tasks, such as Flamingo \cite{alayrac2022flamingo} which trains 10B parameters to adapt a frozen 70B parameter language model, and other successful efficient techniques that are based on vision-conditioned prompt tuning (Frozen \cite{tsimpoukelli2021multimodalfrozen}, PromptFuse \cite{liang2022modularpromptfuse}, LiMBeR \cite{merullo2022linearlylimber}) and adapters (MAGMA \cite{eichenberg2021magma}). This work has demonstrated good performance, showing that it is possible to devise very efficient approaches to adapt existing language models  \cite{gptj, hoffmann2022trainingchinchilla}. On the video side, little work has been proposed, mostly based on Adapters \cite{yang2022zerofrozenbilm, shimomoto2022towards}.  The closest to our approach is PromptFuse \cite{liang2022modularpromptfuse} which finetunes directly for VQA, however, they use encoder-decoder language models and train a soft prompt that is prepended to the input.

\paragraph{Efficient Learning.} Parameter-Efficient learning is an interesting line of research that consists of adapting pretrained models using very few trainable parameters. Prompt Tuning \cite{lester2021power} is one such approach that appends a few learnable tokens, or Soft Prompts to contextualize the input and steer the output of the frozen model toward the desired task. Other approaches use Adapters \cite{houlsby2019parameter, bapna2019adapter}, which are trainable MLP, consisting of 2 linear projection layers with activation in between and inserted inside the model to adapt the self-attention and feedforward layers. Many other approaches have been proposed in the context of NLP such as LoRa \cite{hu2022lora}, Bitfit \cite{zaken2022bitfit}, Hyperformer \cite{mahabadi2021parameter}, Compacters \cite{karimi2021compacter} and (IA)$^3$ \cite{liu2022fewia3}. These approaches have been successfully adapted to other modalities such as image \cite{vpt, chen2022adaptformer}, image-text \cite{sung2022vl, sunglst, zhou2022conditional}, with very little work on video \cite{pan2022stadapter} and Audio \cite{kim2022integrated}. 

Another line of research is Data-Efficient techniques, where the objective is to attain similar performance by significantly reducing the training datasets. Recently, some efforts have been proposed for vision \cite{touvron2021trainingdeit}, language \cite{du2022glam} and vision-language \cite{vicha, cheng2022vindlu, chen2022visualgpt}, which mostly focus on designing better training objectives \cite{vicha}. However, little work has been done to investigate the connection between parameter efficiency and data efficiency, which is considered in this work.

\section{Framework}
\label{sec:method}

To solve multimodal tasks, we propose to augment pretrained LLMs with perception through unimodal perceptual encoders (Fig.\ref{fig:teaser}). We detail our approach in the following.

\begin{figure}[h]
    \centering
    \resizebox{\linewidth}{!}{
    \includegraphics[width=\linewidth]{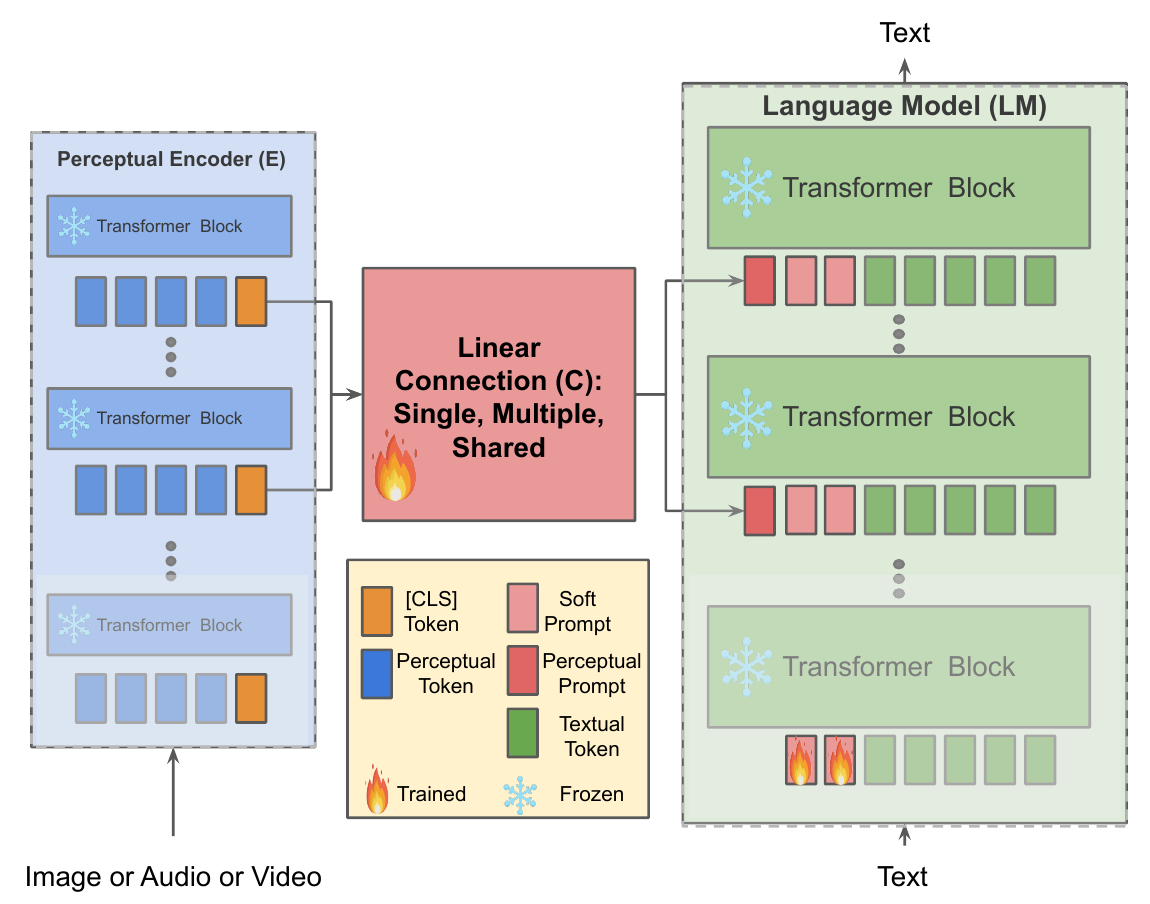}
    }
    \caption{Illustration of the adaptation mechanism in eP-ALM. The perceptual input (image/video/audio) is fed to the perceptual encoder $\textcolor{vis}{\bf{E}}$ (\emph{e.g.}, ViT) and the corresponding text to the $\textcolor{lm}{\bf{LM}}$ (\emph{e.g.}, OPT), which then generates a text conditioned on the perceptual input. The multimodal interaction is done via the $\textcolor{cls}{[CLS]}$ tokens acting as Perceptual Prompt, and are extracted from the last layers of the encoder, then injected in the last layers of $\textcolor{lm}{\bf{LM}}$, after passing by the Linear Connection $\textcolor{con}{\bf{C}}$. The previous $\textcolor{cls}{[CLS]}$ token is replaced by the new one coming from a deeper layer, keeping the number of tokens fixed. The first layers (grayed) of each model are kept intact without any modality interaction. We ease the adaptation with a \textcolor{pink}{Soft Prompt} that is prepended to the input of $\textcolor{lm}{\bf{LM}}$.
    }
    \label{fig:main_v2}
\end{figure}

\subsection{eP-ALM}
\label{sec:epalm_method}

We augment a pretrained LM with perception through several modality-specific encoders. The encoders interact with LM through linearly projected, modality-specific [CLS] tokens. To ease the adaptation, we leverage some parameter-efficient techniques, such as Prompt Tuning. In this section, we detail the design principles of our approach, which is illustrated in Fig.\ref{fig:main_v2}. 

\paragraph{Language Model (LM)}
We adopt OPT models \cite{zhang2022opt}, which are autoregressive language decoders consisting of Self-Attention and Feed Forward layers. They are trained with \textit{next token prediction} objective on 180B tokens mostly in English and gathered from different datasets \cite{gao2020pile, baumgartner2020pushshift}. The authors released a family of models with different scales, starting from 125M up to 175B model size. Besides being open source and trained on English data, the different model sizes allow us to readily investigate the effect of scale, and help to devise new approaches with affordable model sizes.

\paragraph{Perceptual Encoders}
We favor only unimodal models. For images, we use the vanilla ViT model \cite{vit} which consists of Self Attention and FeedForward layers and is pretrained for image classification on ImageNet \cite{russakovsky2015imagenet}. For Video, we use TimeSformer \cite{bertasius2021spaceTimeSformer} that consists of a ViT-like model augmented with temporal attention and pretrained on kinetics \cite{carreira2018short}. For Audio, we adopt AST \cite{gong21b_interspeech_ast}, a vanilla adaptation of ViT to digest spectrograms, that is pretrained on AudioSet \cite{audioset}. Even though we consider only these 3 encoders, the extension of the approach to other types of encoders and modalities is straightforward.

\paragraph{Perceptual Prompt Injection.}  LMs are usually controlled via different textual prompts, such as questions and instructions. Here, the LM is controlled by both the text and the perceptual encoders. Specifically, the projected perceptual tokens are prepended to the textual tokens. Naively using all visual tokens, adds significant computation costs during training and inference, due to the quadratic complexity of attention layers with the number of tokens. This becomes more apparent with LLMs. To mitigate this, we consider only the {[CLS]} token of the perceptual encoders and prepend it to the text tokens. This increases the total number of tokens by 1 which maintains almost the same inference speed.

\paragraph{Connecting Models with Cross-Modal Hierarchical Linear layers.} When freezing the perceptual encoders and language models, the minimal number of trainable parameters are those that amount to connecting these two models while adjusting the embedding dimensions in case of a mismatch. Therefore, we base our approach on this constraint and train only one linear projection layer (single connection, Fig.\ref{fig:main_v2}) to connect both models. To exploit the hierarchical representation encoded in pretrained models, instead of taking only the [CLS] token of the last output layer, we take the [CLS] tokens from several layers of the perceptual model, and we inject these tokens into several layers of the LM (shared connection). The tokens coming from early layers are injected earlier and are then replaced by those coming from deeper layers. 
We favor only the deeper layers (\emph{e.g.}, the last 6 layers of the ViT-B/16, and the last 12 layers of OPT-350M) where the representations are more abstract and less modality-specific. Moreover, using the same linear projection at different representation levels might not help to capture the particularity of such a hierarchy, to this end, we also experiment with different linear layers for each level (multiple connections).

\paragraph{Multimodal Adaptation with Parameter-Efficient Techniques.}
We explore several parameter-efficient techniques to ease the adaptation to multimodal tasks. The main technique we use is \textit{Prompt Tuning} \cite{lester2021power}: it consists of prepending trainable tokens or Soft Prompts to the textual tokens input of the LM. This gives useful context to steer the model output. Contrary to hard prompts that are manually engineered, this provides a more flexible and easier approach for task-dependant contextualization. For the sake of efficiency, we prepend only 10 learnable tokens. We also experiment \textit{Adapters} \cite{houlsby2019parameter} as detailed later.
The approach can be formalized as follows (better read with Fig.\ref{fig:main_v2}):
\vspace{-0.2cm}
\begin{align}
 \textcolor{cls}{[CLS]_{i}} & = \textcolor{con}{\bf{C}}(\textcolor{vis}{\bf{E}_i}(X)),  &i=N_{E}/2, ..., N_{E}, \nonumber \\
\textcolor{lm}{t_{j}} \quad & = \textcolor{lm}{\bf{LM}_j}(\textcolor{cls}{[CLS]_{i}}, \textcolor{pink}{p_{j-1}}, \textcolor{lm}{t_{j-1}}),  &j=N_{L}/2,... , N_{L}, 
\end{align}
where $\textcolor{cls}{[CLS]_{i}}$ is the perceptual token of the input $X$ extracted from the layer i of the perceptual encoder ($\textcolor{vis}{\bf{E}_{i}}$) with $N_{E}$ layers.  $\textcolor{cls}{[CLS]_{i}}$ is projected using the linear connection $\textcolor{con}{\bf{C}}$ and prepended, alongside the Soft Prompt $\textcolor{pink}{p}$ to the embeddings of the textual tokens $\textcolor{lm}{t_{j-1}}$ coming from previous layer in the LM ($\textcolor{lm}{\bf{LM}_{j-1}}$). This operation is repeated each 2 layers in the LM (with $N_{L}$ layers).

\subsection{Efficiency-driven Training Framework Setup}

Current approaches still rely on many costly components that hinder their adoption by the large community. Specifically; they \textbf{(1)} still train a lot of parameters (\emph{e.g.} vision encoders \cite{tsimpoukelli2021multimodalfrozen} and adapters \cite{eichenberg2021magma} with $\sim$325M params/5.11\%), \textbf{(2)} still maintain the multimodal pretraining with image-text pair datasets on top of the unimodal pretraining \cite{merullo2022linearlylimber, eichenberg2021magma, tsimpoukelli2021multimodalfrozen}, \textbf{(3)} leverage multimodal encoders such as CLIP, pretrained on 400M image-text pairs \cite{eichenberg2021magma, merullo2022linearlylimber}, \textbf{(4)} add significant computation overhead during inference, due to the long visual prompt, especially when evaluating with In Context Learning (ICL), that becomes common with LLMs \cite{eichenberg2021magma, merullo2022linearlylimber}.
In this work, we propose a new setup to adapt unimodal models for multimodal downstream tasks. The setup is more challenging and is motivated by the quest for the least effort needed to exploit pretrained models.  The setup is the following:

\begin{itemize}
\setlength\itemsep{0.5pt}
    \item Training only adaptation parameters (\emph{e.g.}, Soft Prompt, linear connection), while keeping as much as possible of pretrained parameters frozen (parameter efficient). 
    \item Avoiding multimodal pretraining and finetuning directly on downstream multimodal datasets (data/compute efficient).
    \item Using only pretrained unimodal models, and avoid using multimodal encoders pretrained on huge datasets (data efficient).
    \item Keeping fast inference (\emph{e.g.}, 1 additional token), by avoiding long prompts, and using additional heavy modules (compute efficient).
    \item Using decoder-only language models (\emph{e.g.}, OPT), the current architecture adopted by LLMs (due to its pretraining efficiency and open-ended generation capacity).
\end{itemize}

Specifically, we train only the linear connection and the soft prompt directly on the downstream multimodal tasks. This amounts to less than 1\% of trainable parameters that we can push further to 0.06\% with big models.

\paragraph{The Pretrain Zero-shot Setup.}
The focus of this work is direct finetuning on target datasets. However, the proposed mechanism (Sec.\ref{sec:epalm_method}) can be adapted straightforwardly to the pretrain-zeroshot setup. In the appendix, we show that eP-ALM outperforms previous work and it is competitive with recent SoTA following the zero-shot evaluation. 

\section{Experiments}
\label{sec:results}

\paragraph{Implementation details.} We use OPT-2.7B in our main model, eP-ALM, and we experiment in Section~\ref{sec:ablation} with OPT models of various sizes.
We extract the [CLS] tokens of the last 6 layers of perceptual encoders and prepend them, after a linear projection, to the text tokens of the last 12 layers of the OPT. Note that we replace the previous [CLS] with the new one to keep the same number of tokens.

For VQA and VideoQA, we cast the problem as open-ended generation and compute the accuracy after a strict comparison between the output text (without truncation) and the ground truth one. Note that this setting is more challenging compared to classification-based VQA and not in favor of our approach as the model might generate semantically correct answers but using different words. We use a special token (`\verb|</a>|') to separate the question from the answer. For captioning, we report the widely adopted CIDEr and  BLUE@4 scores. 
We finetune with the classical cross-entropy loss used to train the original OPT for VQA and Captioning tasks. We use the AdamW optimizer with a learning rate (lr) of 1e-5 warmed up to 2e-5 then decreased to 1e-6 using a cosine scheduler. 
We train for 8 epochs with a batch size of 64 (128 for GQA) and an image resolution of 224. Training our approach with OPT-2.7B for VQA v2 can be done on a single V100 GPU 32GB for few hours. More details are given in the appendix. 
We find the method sensitive to the text decoding approach (Tab. \ref{tab:decoding}). Following other work, we use greedy decoding with beam search for the main results (Sec. \ref{sec:main_results}), and multinomial/random sampling for the ablation study (Sec. \ref{sec:ablation}).

\paragraph{eP-ALM Variants.} Our main model, \textbf{eP-ALM} (illustrated in Figure~\ref{fig:main_v2}), has multiple linear connections; specific learned linear layers for each [CLS] token injected in the model. In addition to Prompt Tuning. 
We also test variants of this model: \textbf{eP-ALM$_{ada}$} (eP-ALM with Adapters instead of Soft Prompts), \textbf{eP-ALM$_{lin}$} (trains a shared linear connection with all [CLS] tokens, and no prompt tuning) and \textbf{eP-ALM$_{pt}$} ($lin$ + Soft Prompt). For Adapters, we follow other work \cite{eichenberg2021magma} and add sequentially one adapter module after self-attention and feedforward layers in all the blocks of OPT. While this might give better results, it adds a significant number of trainable parameters.

\subsection{Main Results}
\label{sec:main_results}
In this section, we present the main comparison with other approaches. We compare eP-ALM to SoTA in Section~\ref{sec:sota}, then present more detailed results for the image modality in Section~\ref{sec:img-text}, the video modality in Section~\ref{sec:video-text}, and the audio modality in Section~ \ref{sec:audio-text}.

\subsubsection{Comparison with SoTA}
\label{sec:sota}
We start by comparing eP-ALM to other SoTA that trains large number of parameters and most often with large-scale pretraining. Tab. \ref{tab:sota} shows a comparison with both zero-shot (ZS) and Finetuning (FT) setups. The performance of eP-ALM is generally higher than ZS scores and still below FT ones. However, the performance gap with FT models, is smaller with the audio and video modalities.

\begin{table*}[h]
    \small
    \centering
    \setlength\tabcolsep{4pt}
    \resizebox{0.98\linewidth}{!}{
        \begin{tabular}{lcc@{\hspace{25pt}}c@{\hspace{25pt}}cc@{\hspace{25pt}}c@{\hspace{25pt}}c@{\hspace{25pt}}cc@{\hspace{25pt}}c}
            \toprule
            & VQA v2 & GQA & \multicolumn{2}{c}{COCO} & MSVD-QA & MSRVTT-QA & \multicolumn{2}{c}{MSR-VTT} & \multicolumn{2}{c}{Audiocaps} \\
            & Test/Val std & Test & CIDEr & B@4 & Test & Test & CIDEr & B@4 & CIDEr & B@4 \\
            \midrule
            SoTA (PT+ZS) & 56.30$^*$/-- \cite{alayrac2022flamingo} & 29.3 \cite{jin-etal-2022-good_fewvlm} & 84.3 \cite{alayrac2022flamingo} & -- & -- & 17.4 \cite{alayrac2022flamingo} & -- & -- & -- & -- \\
            SoTA (PT+FT) & 84.3$^*$/-- \cite{chen2022pali} & 60.8 \cite{cho2021unifyingvlt5} & 145.3 \cite{wang2022unifyingofa} & 44.9 \cite{wang2022unifyingofa} & 56.6 \cite{li2022lavender} & 45.0 \cite{li2022lavender} & 60 \cite{seo2022endmvgpt} & 48.92 \cite{seo2022endmvgpt} & 75.3 \cite{gontier_audiotags} & 26.6 \cite{gontier_audiotags} \\
            \midrule
            eP-ALM (FT) & 54.89/54.78 & 42.91 & 111.63 & 33.35 & 38.40 & 35.90 & 48.79 & 38.51 & 61.86 & 20.81 \\
            \bottomrule
        \end{tabular}
    }
    \vspace{2ex}
    \caption{
        \footnotesize
        Comparison of eP-ALM$_{pt}$-L (greedy decoding with number of beams 3) to SoTA that train large number of parameters on large datasets. $^*$: test-std set. PT: pretrained, FT: finetuned, ZS: zero-shot.
    }
    \label{tab:sota}
\end{table*}

\subsubsection{Image-Text Results}
\label{sec:img-text}
We use a frozen ViT-B/16 pretrained on ImageNet1K as the image encoder.
We consider the following image-text benchmarks; VQA v2 \cite{goyal2017makingvqav2}, GQA \cite{hudson2019gqa} and COCO Caption \cite{chen2015microsoftcaption}. We use Karpathy splits for VQA v2 and COCO, unless specified otherwise. For the following sections we use gready-decoding with beam search (number of beams=1)

\paragraph{Baselines.} As we are the first to propose this setup, to have a fair comparison, we reimplemented some of the existing approaches and use the same vision (ViT-ImageNet) and language (OPT) models for all:

\noindent\textit{1) B$_{PromptFuse}$}; which is equivalent to PromptFuse \cite{liang2022modularpromptfuse} and uses Prompt Tuning (N=10). We add a linear projection for the last [CLS] token. The [CLS] token is prepended to the input of the LM. Note that we could not avoid adding a trained linear projection as there is a mismatch between the dimensions of the vision and language model. 

\noindent\textit{2) B$_{MAGMA}$}; which is equivalent to MAGMA \cite{eichenberg2021magma} and uses Adapters. We prepend the [CLS] token to the input of LM after linear projection. Note that, we consider only the [CLS] token as we find it better than prepending all image tokens (eP-ALM$_{MAGMA}^*$). We also find that training the ViT degrades the performance, thus we keep it frozen in favor of their approach. 

\noindent\textit{3) B$_{LimBEr}$}; which is equivalent to LimBEr \cite{merullo2022linearlylimber} and only trains the linear projection to project visual tokens and prepend them to the input text. Similarly, we only consider the [CLS] token as it gives better accuracy.

\paragraph{Comparison to Other Work.} Based on our study (Sec. \ref{sec:ablation}), we use ViT-B/16 and OPT-2.7B in our main model and in our replication of other approaches.
In Table \ref{tab:comp_2_7b} we compare with other work on VQA v2, GQA, and COCO Caption. We significantly outperform other approaches with at least +10 points on VQA v2, +9 points on GQA and we double the scores on COCO Caption. eP-ALM$_{pt}$-L with OPT-6.7B and ViT-L gives the best scores while training only 0.06\% of model parameters.

Note that for COCO Caption, other works give very low scores (thus we did not report them).

\begin{table}[h]
    \small
    \centering  
 \setlength\tabcolsep{4pt}
    \resizebox{0.9\linewidth}{!}{%
    \begin{tabular} {lcc@{\hspace{24pt}}cc@{\hspace{24pt}}cc}
        \toprule        
       \multirow{2}{*}{Method}  & \multicolumn{2}{c@{\hspace{24pt}}}{VQA v2}  & \multicolumn{2}{c@{\hspace{24pt}}}{GQA} & \multicolumn{2}{c}{COCO} \\
  \cmidrule(lr{24pt}){2-3} \cmidrule(lr{24pt}){4-5} \cmidrule(lr){6-7}
     &  Val & Test & Val & Test & B@4 &  CIDEr \\
  \midrule
   PromptFuse$^\dagger$ \cite{liang2022modularpromptfuse}     & 34.1$^\dagger$  & -- & -- & -- & -- & -- \\
     \midrule
 B$_{LimBEr}$     &  34.1 &  33.5 & 30.81  & 29.4 & -- & --   \\
 B$_{PromptFuse}$ & 40.4  &  39.5 & 33.74  & 31.51 &  15.05 &  48.26   \\
 B$_{MAGMA}$      & 32.2  & 31.8  & 30.98  &  28.93  & -- & --   \\
 \midrule 
  eP-ALM$_{pt}$ & 48.8 & 47.8 & 43.8 & 40.3  &  27.52 &  91.92   \\
  eP-ALM & \textbf{50.7}/\textbf{53.3}$^\dagger$  & \textbf{50.2} & \textbf{45.0} & \textbf{40.4} & \textbf{29.47} & \textbf{97.22}   \\
  \midrule
    eP-ALM$_{pt}$-L$^*$ &  54.58/54.47$^\dagger$  & 54.47 & 46.86  &  42.7 & 31.24 &  107.0     \\ 
 \bottomrule
    \end{tabular}
    }
    \vspace{2ex}
    \caption
    {
    \footnotesize   
        Comparison with other work after direct finetuning on VQA v2, GQA, and COCO Caption. eP-ALM significantly outperforms other approaches. eP-ALM uses ViT-B/16 and OPT-2.7B. eP-ALM-L uses OPT-6.7B and  ViT-L/16.
        $\dagger$: use standard split.$^*$: trained more than 8 epochs.
        }
    \label{tab:comp_2_7b}

\end{table}

\paragraph{Few-shot Results: Are Parameter-Efficient Models also Data-Efficient?}
In this section, we investigate how data-efficient our model can be. To this end, we train on a very small portion (randomly sampled) from the VQA training set and evaluate on the validation set.  Table \ref{tab:fewshot}, shows the superiority of our approach over other baselines. Interestingly, we can achieve 80\% (41.9 vs 52.77) of the performance when training on 1\% of the data. This validates the approach on low resources scenarios and shows that, in addition to being parameter-efficient, our model is also data-efficient.

\begin{table}[h]
    \small
    \centering  
 \setlength\tabcolsep{4pt}
    \resizebox{0.7\linewidth}{!}{%
    \begin{tabular} {lc@{\hspace{24pt}}c}
        \toprule        
     Method  & Train. data \% (\# of shots) & VQA v2  \\
  \midrule
 PromptFuse$^*$ \cite{liang2022modularpromptfuse}   & 0.12\% (512) & 29.40 \\    
 \midrule
  B$_{LimBEr}$  & 1\% (4.4K) &   28.9   \\ 
  B$_{PromptFuse}$  & 1\% (4.4K) &  31.9    \\ 
  B$_{MAGMA}$  & 1\% (4.4K) &  34.5    \\ 
 \midrule 
  eP-ALM$_{lin}$$^*$ & 0.12\% (512) &  31.3 \\
  eP-ALM$_{pt}$ & 0.12\% (512) &   30.36 \\    
  eP-ALM & 0.12\% (512) &  35.54 \\ 
  eP-ALM & 1\% (4.4K) &   41.9   \\ 
  eP-ALM & 10\% (44K) &   47.4   \\ 
  eP-ALM & 100\% (443K) & \textbf{52.77} \\ 
    \bottomrule
    \end{tabular}
    }
    \vspace{2ex}
    \caption
    {
    \footnotesize   
        Few-shot Results on VQA v2 validation set (standard split). $^*$: longer training.
        }
    \label{tab:fewshot}

\end{table}

\paragraph{Out of Distribution (OOD) Generalization: Do Parameter-Efficient Models Generalize Better?} Here we investigate whether our parameter-efficient approach can perform well in OOD scenarios. To this end, we follow other approaches \cite{agrawal2022rethinking} and train our model on the training set of a given benchmark, and evaluate it on the validation set of another benchmark, without multimodal pretraining. We measure the performance gap, i.e. the accuracy difference between a model trained on a different benchmark and the same model trained on the target benchmark.
Tab.\ref{tab:ood} shows that eP-ALM, that trains 0.06\% of total parameters, is very competitive in terms of OOD accuracy with other baselines, that train all model parameters and pretrain on large amount of data. Specifically, we outperform VILBERT on VQAv2 by more than 2 points.  Interestingly, the OOD-IID gap for eP-ALM, is at least 2 times lower compared to ALBEF \cite{li2021alignalbef} and VilBERT \cite{lu2019vilbert}. This reveals that our parameter-efficient approach generalizes relatively well in OOD scenarios.

\begin{table}[h]
    \small
    \centering  
 \setlength\tabcolsep{4pt}
    \resizebox{0.95\linewidth}{!}{%
    \begin{tabular} {l c@{\hspace{5pt}}c@{\hspace{25pt}}c@{\hspace{5pt}}cc@{\hspace{12pt}}c}
        \toprule        
     \multirow{2}{*}{Method}  & Multimodal & Trained & \multirow{2}{*}{Train data}  & \multicolumn{2}{c@{\hspace{24pt}}}{Test data} & Gap  \\ \cmidrule(lr{24pt}){5-6}
  & PT data & param. (\%) & & VQA v2 & GQA \\
  \midrule

\multirow{2}{*}{ALBEF \cite{agrawal2022rethinking}}  & \multirow{2}{*}{14M} & \multirow{2}{*}{100\%}  & VQA v2 & --\cmmnt{72.1} & 50.1 & -21.8  \\   
           & &                  & GQA    & 50.3 & --\cmmnt{64.2} & -14.1 \\  \midrule 
 \multirow{2}{*}{VILBERT \cite{agrawal2022rethinking}} & \multirow{2}{*}{3M} & \multirow{2}{*}{100\%}    & VQA v2 & --\cmmnt{62.2} & 42.6 & -20.4 \\   
           & &                  & GQA    & 41.8 & --\cmmnt{65.3} & -22.7 \\    
           
 \midrule
  \multirow{2}{*}{eP-ALM$_{pt}$-L} \cmmnt{$^*$(nb3)} & \multirow{2}{*}{0} & \multirow{2}{*}{0.06\%}   & VQA v2 & --\cmmnt{54.78} &  41.39 & \textbf{-9.59} \\   
            & &             & GQA    &  45.19  & --\cmmnt{47.19} & \textbf{-5.8} \\

    \bottomrule
    \end{tabular}
    }
    \vspace{2ex}
    \caption
    {
    \footnotesize   
        Out-Of-Distribution Generalization on GQA and VQA v2 (standard split). The Gap shows the performance degradation when the model is trained on a different dataset. 
    }
    \label{tab:ood}

\end{table}

\subsubsection{Video-Text Results}
\label{sec:video-text}
We investigate how much our approach generalizes to other modalities.
To this end, we evaluate eP-ALM for Video QA on MSRVTT-QA \cite{msvd_msrvtt} and MSVD-QA \cite{msvd_msrvtt} and for Video Captioning on MSR-VTT \cite{Xu_2016_CVPR_msrvtt}. For the video encoding, we use the TimeSformer-base\cite{bertasius2021spaceTimeSformer} model pretrained on Kinetics-600 \cite{k600}. We use 8 and 16 224x224 frames for VQA and captioning respectively. 

\paragraph{Comparison to other work} to the best of our knowledge, FrozenBiLM \cite{yang2022zerofrozenbilm} is the only parameter-efficient work proposing to adapt LMs for video-language tasks. It uses Adapters to adapt the frozen CLIP-ViT and Bidirectional LM for Video QA. We compare our approach to our re-implementation of this baseline; where we train only the Adapters and the linear projection layer to project the last [CLS] token and prepend it to the input text ones. 
The results in Tab. \ref{tab:comp_2_7b_vid} show that eP-ALM outperforms this baseline by a significant margin. The reason why the latter does not give good results might be due to prepending the visual tokens to the input of OPT.
We can reduce the number of parameters and slightly degrade the performance by using a shared linear connection (eP-ALM vs eP-ALM$_{pt}$).

\begin{table}[h]
    \small
    \centering  
 \setlength\tabcolsep{4pt}
    \resizebox{0.95\linewidth}{!}{%
    \begin{tabular} {lc@{\hspace{25pt}}c@{\hspace{10pt}}c@{\hspace{10pt}}c  c }
        \toprule        
     \multirow{2}{*}{Method} & Trained & MSVD-QA  & MSRVTT-QA & \multicolumn{2}{c@{\hspace{10pt}}}{MSRVTT} \\ \cmidrule(lr{10pt}){5-6}
     & param (\%) & Test & Test & CIDEr & B@4  \\
  \midrule

 B$_{FrozenBiLM}$ (from \cite{yang2022zerofrozenbilm}) & 3.72 \% &  14.58  &  6.33  & -   & -      \\ \midrule
  eP-ALM & 0.89 \% & \textbf{38.64} & \textbf{36.16} & \textbf{47.31} & \textbf{38.51}   \\
  eP-ALM$_{pt}$ & 0.54 \% & 38.79 & 35.62 &  45.30 & 39.34  \\

    \bottomrule
    \end{tabular}
    }
    \vspace{2ex}
    \caption
    {
    \footnotesize   
        Comparison with different approaches after direct finetuning on MSVD-QA, MSRVTT-QA, and MSRVTT Caption.
        }
    \label{tab:comp_2_7b_vid}

\end{table}

\paragraph{Zero-Shot Results}
To explore the generalization of our approach, we evaluate on Zero-Shot for VideoQA, where the model is trained on a dataset different from the target one. Table \ref{tab:zs} shows a comparison with other approaches. eP-ALM, trained on VQA v2 (standard split), outperforms other approaches trained on significantly more data. Specifically, eP-ALM outperforms Flamingo-3B \cite{alayrac2022flamingo} on MSRVTT-QA by more than 2 points, and attains double the scores of FrozenBiLM \cite{yang2022zerofrozenbilm}. Contrary to some of other approaches that cast the task as classification (similarity-based) \cite{yang2021justask} or constrained generation through masking, considering only a subset of answers (1k or 2k) \cite{yang2022zerofrozenbilm, zellers2022merlot, li2022lavender}, our approach is evaluated (with a character-wise comparison with the ground-truth) with unconstrained Open-ended Generation (OE Gen) and can generate answers with arbitrary lengths. This is more challenging and not in favor of our approach. 

\begin{table}[h]
    \small
    \centering  
 \setlength\tabcolsep{4pt}
    \resizebox{0.95\linewidth}{!}{%
    \begin{tabular} {lc@{\hspace{10pt}}c@{\hspace{10pt}}c@{\hspace{10pt}}c@{\hspace{10pt}}c }
        \toprule        
     Method  & Training data & Train. Param. (\%) & OE Gen & MSRVTT-QA & MSVD-QA  \\
  \midrule
 JustAsk \cite{yang2021justask}   & ActivityNet-QA & 89.6\% & \xmark &  2.7 & -      \\
 JustAsk \cite{yang2021justask}   & HowToVQA69M  & 89.6\%  & \xmark & 2.9 & 7.5      \\
 LAVENDER \cite{li2022lavender}   & WebVid2.5M+CC3M  & 100\% & \xmark &  4.5 & 11.6      \\
 MERLOT Reserve \cite{zellers2022merlot} & YT-Temporal-1B & 100\% & \xmark & 5.8 & - \\
 FrozenBiLM $^\dagger$ \cite{yang2022zerofrozenbilm} & 400M-CLIP + VQA v2 & 2.9\% & \xmark & 6.9 & 12.6 \\
 Flamingo 3B \cite{alayrac2022flamingo} & M3W+ALIGN+VTP  & 40\% & \cmark & 11.0 & \textbf{27.5} \\
 \midrule 

    eP-ALM & VQA v2 & 0.9\%  & \cmark &  13.17 &  24.82  \\
  eP-ALM $^\dagger$  & VQA v2 & 0.9\%  & \cmark &  \textbf{14.54} & 27.09  \\

    \bottomrule
    \end{tabular}
    }
    \vspace{2ex}
    \caption
    {
    \footnotesize   
    Zero-Shot results on Video QA. OE Gen: unconstrained Open-Ended Generation. $\dagger$ evaluated on questions with top 1k answers.}
    \label{tab:zs}

\end{table}

\subsubsection{Audio-Text Results}
\label{sec:audio-text}

We investigate the generalization of our approach to the audio domain. The encoder is AST-base model \cite{gong21b_interspeech_ast} pretrained for classification on AudioSet \cite{audioset}.
We evaluate on AudioCaps dataset \cite{audiocaps}, the largest benchmark for Audio Captioning. We train with mel spectrograms of 128 bins and frequency and time masking with a batch size of 8.  

To the best of our knowledge, no prior work has been proposed to efficiently adapt LM for audio-text tasks, thus we compare with other end-to-end trained SoTA that takes only the audio signal as input. Tab. \ref{tab:audiocaps} shows that our approach is very competitive with previous work, showing the potential of efficient adaptation of LM for the audio modality.

\begin{table}[h]
    \small
    \centering  
 \setlength\tabcolsep{4pt}
    \resizebox{0.95\linewidth}{!}{%
    \begin{tabular} {lc@{\hspace{25pt}}c@{\hspace{10pt}}c@{\hspace{10pt}}c@{\hspace{10pt}}c@{\hspace{10pt}}c@{\hspace{10pt}}c}
        \toprule        
     \multirow{2}{*}{Method} & Trained & \multicolumn{6}{c@{\hspace{10pt}}}{AudioCaps} \\ \cmidrule(lr{10pt}){3-8}
     & param (\%)  & B@1 & B@2 & METEOR & CIDEr & SPICE & SPIDEr \\
  \midrule
Kim et al. \cite{kim-etal-2019-audiocaps} & 100\% & 0.614 & 0.446 & 0.203 & 0.593 & 0.144 & 0.369 \\
 Koizumi et al. \cite{koizumi2020audio} & 100\% & 0.638 & 0.458 & 0.199 & 0.603 & 0.139 & 0.371 \\
 Eren et al. \cite{eren_sememb2020} & 100\% & \textbf{0.710} & 0.490 & \textbf{0.290} & 0.750 & - & - \\
 Xu et al. \cite{xu2021investigating} & 100\% & 0.655 & 0.476 & 0.229 & 0.660 & 0.168 & 0.414 \\ 
 Mei et al. \cite{mei2021audio} & 100\% & 0.647 & 0.488 & 0.222 & 0.679 & 0.160 & 0.420 \\
 Gontier  et al. \cite{gontier_audiotags} & 100\% & 0.699 & \textbf{0.523} & 0.241 & \textbf{0.753} & \textbf{0.176} & \textbf{0.465} \\
 Liu et al. \cite{liu2022leveraging} & 100\% & 0.671 & 0.498 & 0.232 & 0.667 & 0.172 & 0.420 \\ \midrule

    eP-ALM \cmmnt{(nb3)} & 0.90 \% & 66.08 & 47.57 & 22.69 & 63.61 & 16.29 &  -   \\

    \bottomrule
    \end{tabular}
    }
    \vspace{2ex}
    \caption
    {
    \footnotesize   
        Comparison with other work for Audio Captioning on AudioCaps Test set.
        }
    \label{tab:audiocaps}
\end{table}

\subsection{Ablation Study}
\label{sec:ablation}

In this section, we ablate different component of our work.

In the following, we run some ablations for Image-Text tasks, mostly on VQA v2.

\begin{figure}
    \centering
    \resizebox{1\linewidth}{!}{
    \includegraphics[width=\linewidth]{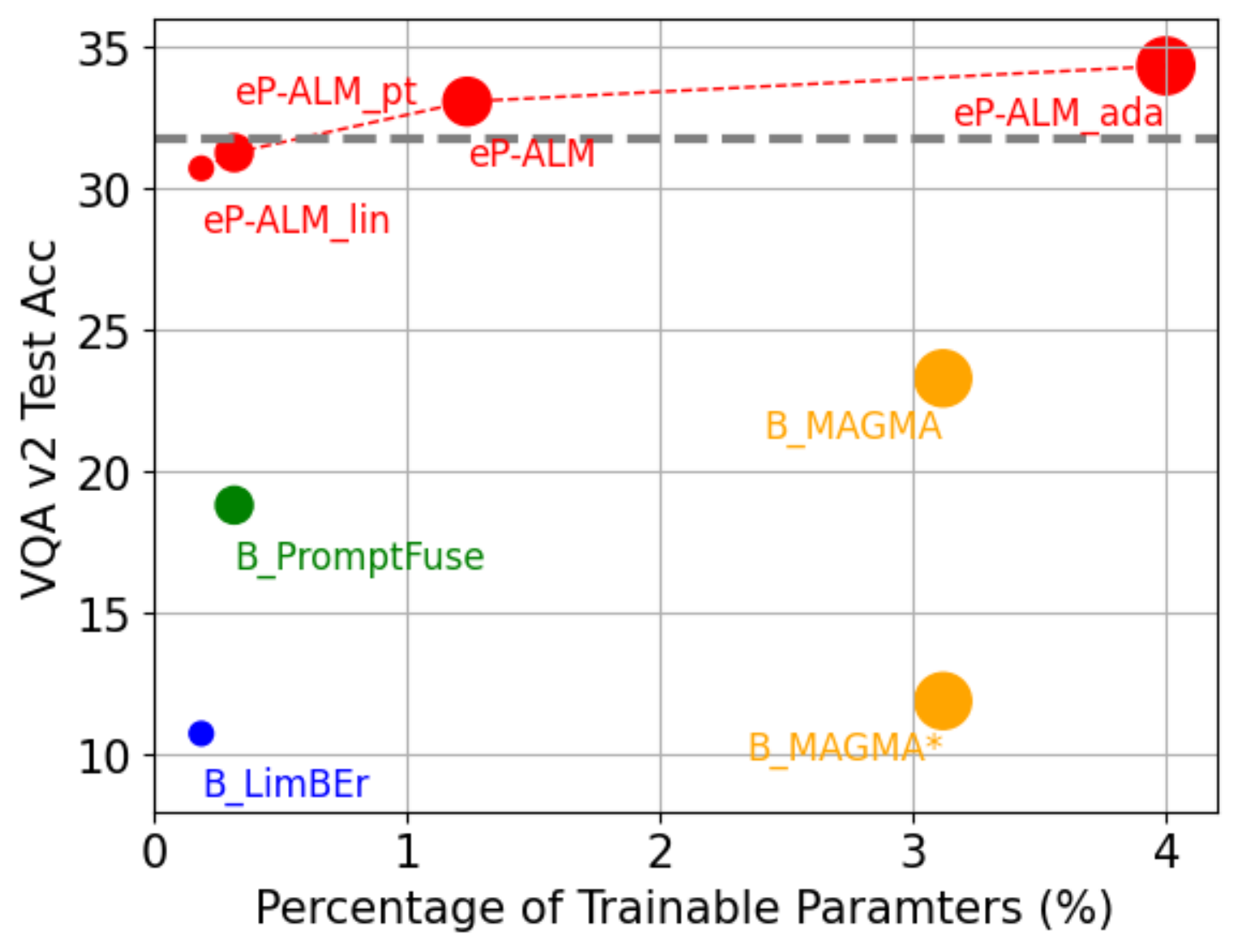}
    }
    \caption{Comparison with other baselines on VQA v2. eP-ALM uses ViT-B/16 and OPT-350M. \textcolor{red}{Our approach} significantly outperforms other approaches. eP-ALM already surpasses the dashed "upper" \textcolor{gray}{baseline} that trains all model parameters. Allocating more adaptation parameters help to increase the scores. Marker size indicates model size.}
    \label{fig:comp_350m}
\end{figure}

\paragraph{Comparison with different variants and baselines.} We start by comparing the different variants to other work in Fig. \ref{fig:comp_350m}. All models use OPT-350M and ViT-B/16. Other approaches lag significantly behind our model. B$_{MAGMA}$ gives the best results (23.3\% acc.) among them, followed by  B$_{PromptFuse}$ (18.82\% acc.) and finally B$_{LimBEr}$ (10.75 \% acc.). We also compare with another MAGMA baseline (B$_{MAGMA}$$^*$) that prepends all visual tokens to the input, and we find a significant degradation compared to passing only the [CLS] token. This reveals that prepending all visual tokens directly to the input hinders the adaptation.

We can notice a consistent improvement of eP-ALM when adding more trainable parameters. The most parameter-efficient model is eP-ALM$_{lin}$ which has 30.72\%, while the best has 34.34\% (with Adapters eP-ALM$_{ada}$). Interestingly, eP-ALM$_{lin}$ with only one linear layer succeeds to get good performance on this challenging setup, revealing that the language and visual representation spaces are not very far. 
Other parameter-efficient techniques such as Prompt Tuning can help to get additional points (30.72 with eP-ALM$_{lin}$ vs 31.27 with eP-ALM$_{pt}$). Moreover, using different layers for each injected [CLS] token seems to give significant improvement (31.27 with eP-ALM$_{pt}$ vs 33.08 with eP-ALM). 

Finally, we show that eP-ALM surpasses the ``full finetuning'' baseline (grayed line) that finetune all parameters by 1.27 points (31.79 vs 33.08). This reveals that training all weights of pretrained models on small datasets can reduce their generalization capability and degrades performance.

As a trade-off between performance and efficiency, we favor eP-ALM which we carry on for the following study.

\paragraph{Extraction and Injection Level of [CLS] Tokens.} Here we investigate which [CLS] tokens to extract from the ViT and where is the best position to inject them inside the OPT model. Table \ref{tab:ab_350m} shows that extracting the last [CLS] tokens (from the last 6 layers) is better than using only the last one, as done in other approaches (Acc 30.53 vs 33.08). In addition, using all [CLS] tokens seems to degrade the performance. Moreover, prepending [CLS] tokens to all OPT layers degrades slightly (33.08 vs 32.15), and prepending to the input of OPT gives the worst results. This might indicate that it is easier to merge visual and textual tokens deeper in the model, where the representations are more abstract, compared to the first layers where we have more modality-specific features and higher representation mismatch.

\begin{table}[!t]
    \small
    \centering  
 \setlength\tabcolsep{4pt}
    \resizebox{0.7\linewidth}{!}{%
    \begin{tabular} {c@{\hspace{10pt}}c@{\hspace{25pt}}c}
        \toprule        
     \multicolumn{2}{c@{\hspace{10pt}}}{[CLS] tokens} &  VQA v2  \\ \cmidrule(lr{10pt}){1-2}
     From ViT layers & To OPT layers & Test Acc. \\
     \midrule
 12   & 12 to 23 & 30.53   \\
 6 to 12 & 12 to 23 & 33.08   \\
 1 to 12 & 12 to 23 & 31.17    \\
 \midrule
 6 to 12 & 1 to 23 & 32.15    \\
 12   & 1 & 18.82   \\

    \bottomrule
    \end{tabular}
    }
    \vspace{2ex}
    \caption
    {
    \footnotesize   
        Ablation study.  Extracting the [CLS] tokens from the last layers of ViT (layers 6 to 12) is better than taking only the last token (layer 12). Injecting the [CLS] tokens lately (layers 12-24) in the OPT is better than injecting them in all layers or only in the input. 
        }
    \label{tab:ab_350m}

\end{table}

\begin{figure}
    \centering
    \resizebox{0.85\linewidth}{!}{
    \includegraphics[width=\linewidth]{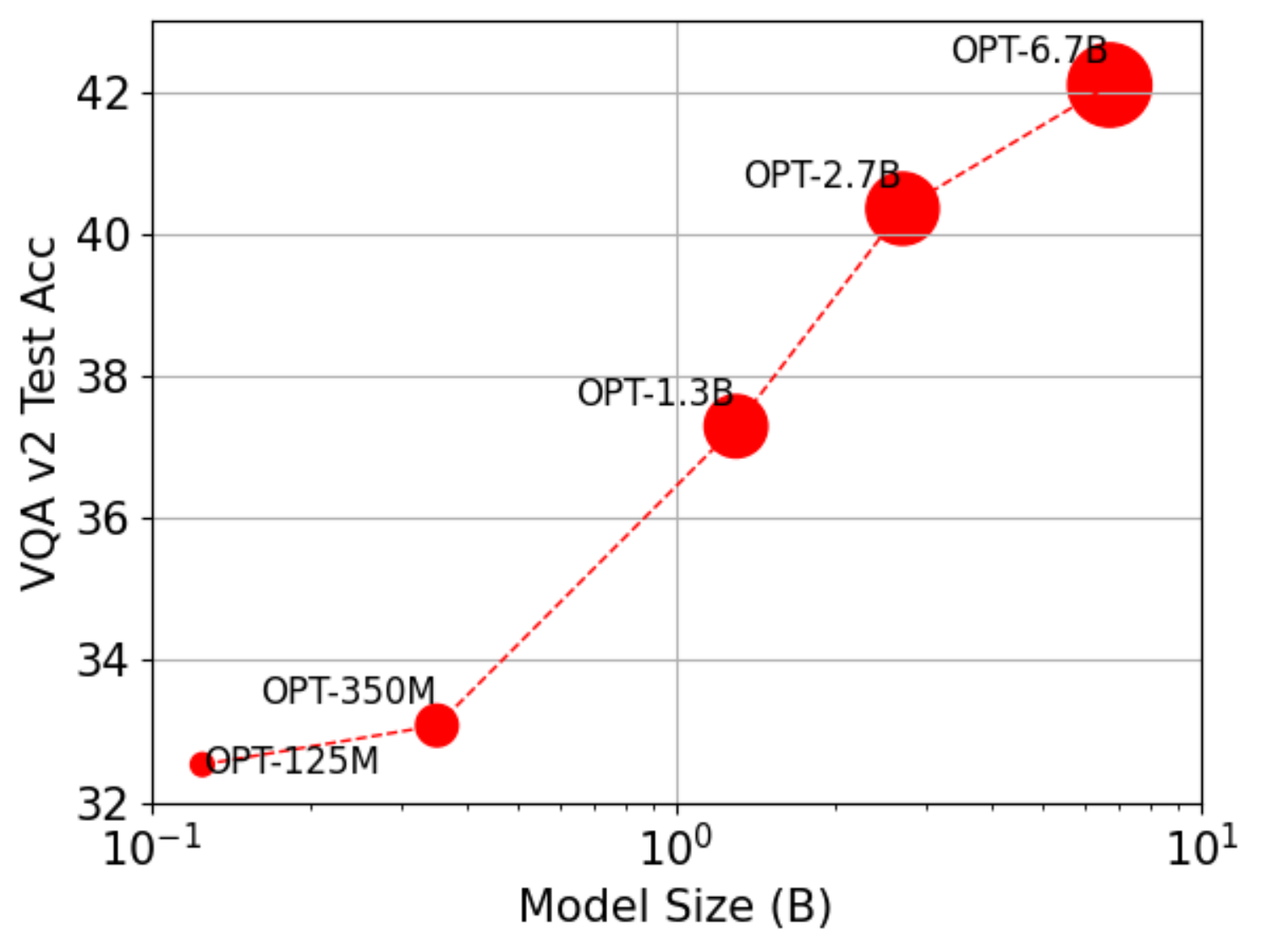}
    }
    \caption{Scaling LM; we scale the OPT from 125M to 6.7B parameter. eP-ALM becomes more effective with scale. The biggest performance jump is when scaling beyond 1B parameters (1.3B).}
    \label{fig:scale}
\end{figure}

\paragraph{Scaling LM.} An interesting question that we investigate is the impact of scaling the language model's parameters on our approach. 
Ideally, we would like to have an approach that efficiently exploits LLMs for other tasks and modalities, without having access to enormous computational resources. In Table \ref{fig:scale}, we show that the scores increase with the model size with the biggest jump being between OPT-350M (33.08 vs 37.29) and OPT-1.3B ($\sim \times 4$  the model size). The consistent improvement with scale shows the effectiveness of the approach when considering very big models.

As a trade-off between performance and model size, we favor OPT-2.7B and use it for all other experiments.

\begin{figure}
    \centering
    \resizebox{0.85\linewidth}{!}{
    \includegraphics[width=\linewidth]{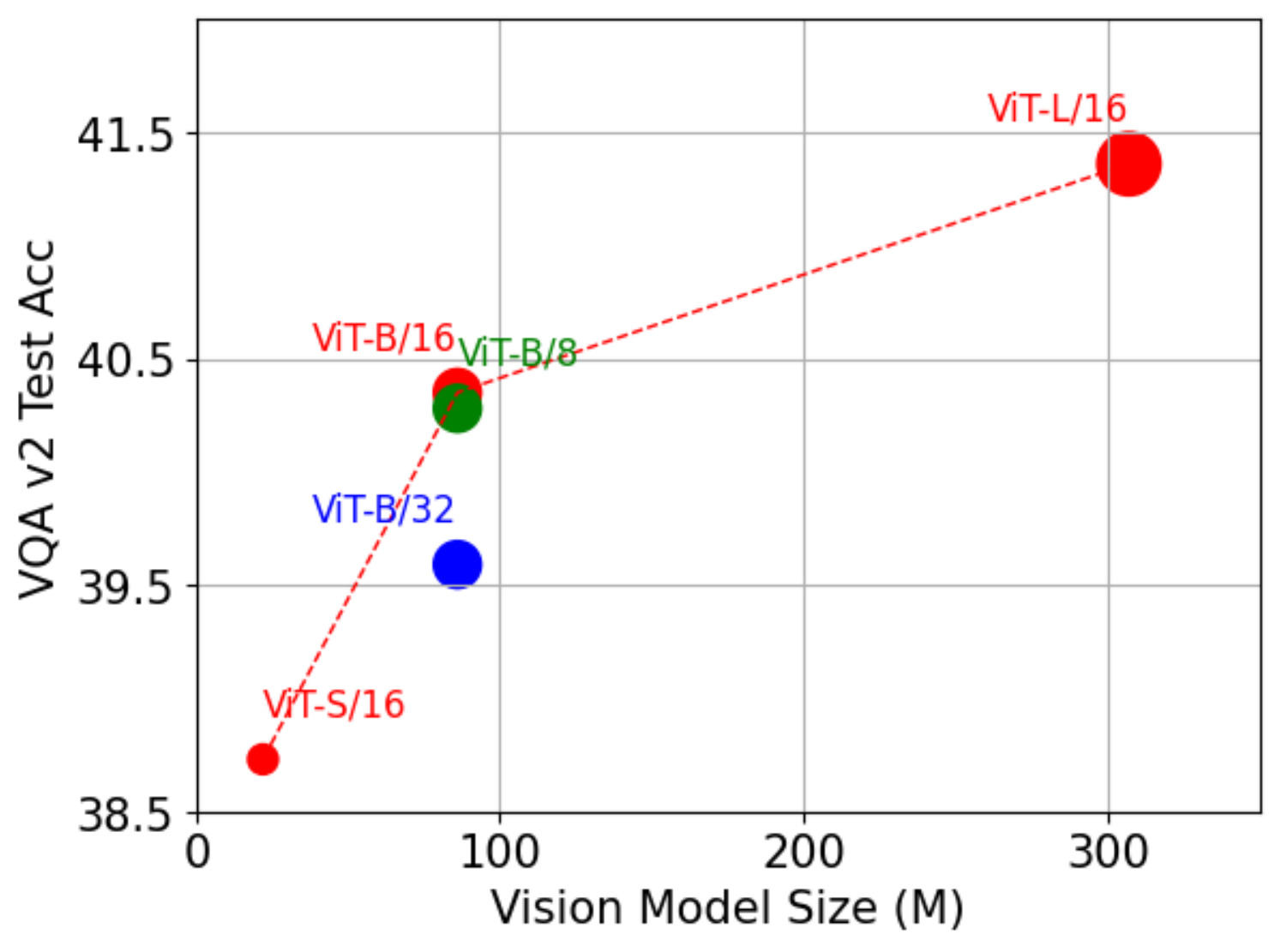}
    }
    \caption{Scaling Vision Model; the score increases with the size of the ViT. Increasing the patch resolution beyond 16 does not help.}
    \label{fig:scale_vis}
\end{figure}

\paragraph{Scaling Visual Model.} We also study how the model behaves when scaling the visual encoder. In Figure \ref{fig:scale_vis}, we can notice that the scores increase with the size of the ViT. The best is ViT-L/16 (41.36) and the worst is ViT-S/16 (Acc 38.73). However, the ViT resolution or the number of image patches/tokens does not seem to have a significant effect on the final performance after a resolution of 16. \\
\paragraph{Scaling Compute.} Table \ref{tab:compute} shows that our approach scales with compute, as training for more epochs leads to 4 points gain in VQA accuracy. Interestingly with OPT-6.7B and ViT-L (eP-ALM$_{pt}$-L), we achieve a score of 43.6 by training only \textbf{0.06\%} of model parameters ($\sim$4M params).

\begin{table}[h]
    \small
    \centering  
 \setlength\tabcolsep{4pt}
    \resizebox{0.5\linewidth}{!}{%
    \begin{tabular} {lcc}
        \toprule        
       \multirow{2}{*}{Method}  & number of & VQA v2   \\
     & epochs & Val   \\
  \midrule
  eP-ALM & 8 & 38.9   \\
  eP-ALM & 32 & 42.9   \\ \midrule
  eP-ALM$_{pt}$-L & 8 & 42.5 \\
  eP-ALM$_{pt}$-L & 32 & 43.6 \\

    \bottomrule
    \end{tabular}
    }
    \vspace{2ex}
    \caption
    {
    \footnotesize   
        Scaling Compute. Evaluation on VQA v2 standard split.
        }
    \label{tab:compute}

\end{table}

\begin{figure}[h]
    \centering
    \resizebox{0.9\linewidth}{!}{
    \includegraphics[width=\linewidth]{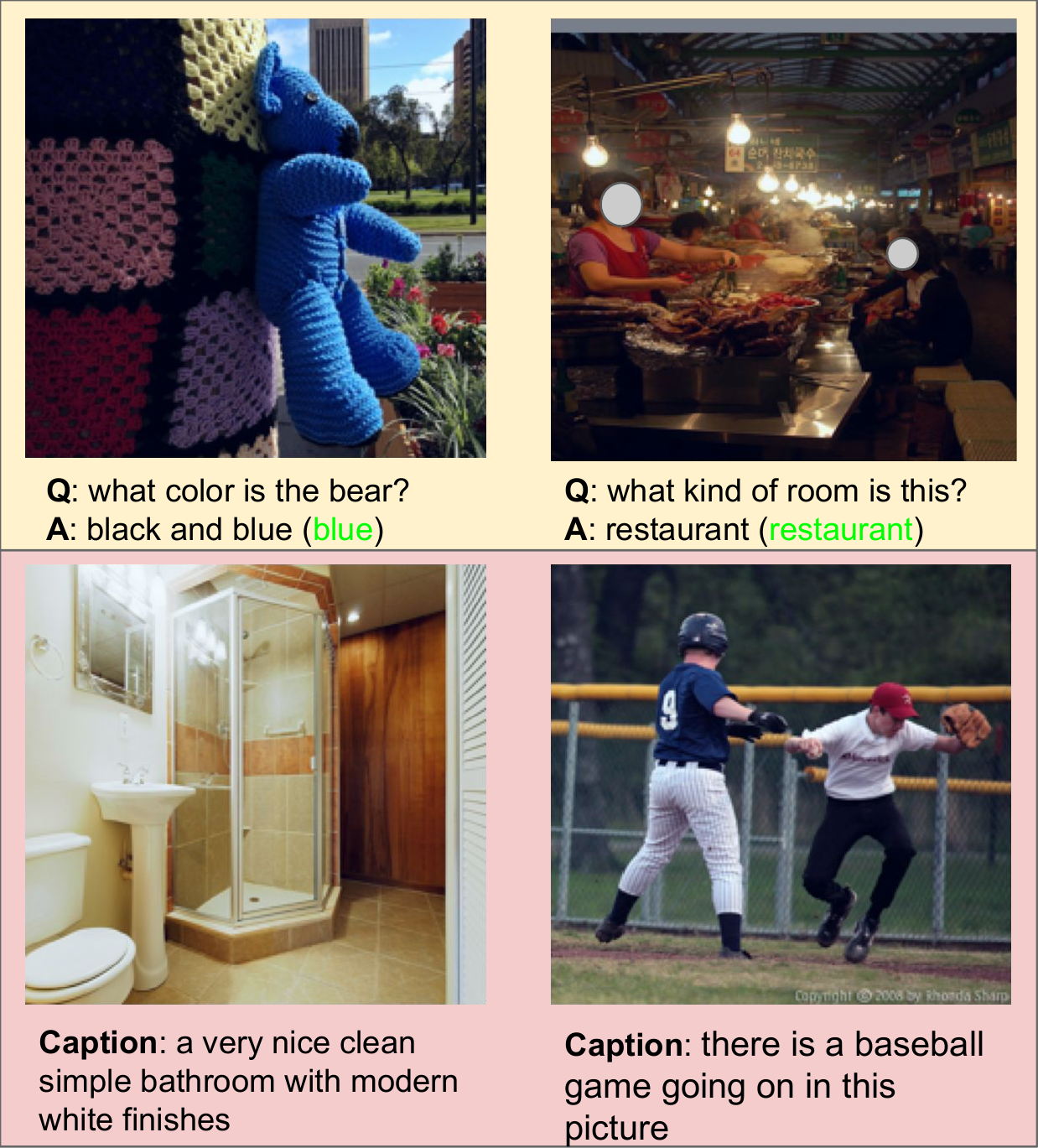}
    }
    \caption{Qualitative results of eP-ALM: the model is able to generate accurate answers and coherent descriptions of the image.}
    \label{fig:qual_mini}
\end{figure}

\paragraph{Qualitative Results.} We show some qualitative results of our eP-ALM model with OPT-2.7B in Fig. \ref{fig:qual_mini}. For VQA, we can notice that our model is able to correctly answer the questions. Moreover, some of the answers are richer and more accurate than the manually labeled ground truth in the dataset. This also reveals that the exact matching evaluation protocol is not in favor of the open-ended generation produced by our model. Interestingly, it seems that the model learned the answering style in the training set (\emph{i.e}, short and concise answers). For Captioning, the model can generate coherent sentences describing the image globally. However, it still misses some details in the image.

\section{Conclusion}
\label{sec:conclusion}

In this work, we propose a new challenging setup to efficiently adapt unimodal models for multimodal tasks, which is centered around augmenting existing LMs with perception. Without multimodal pretraining, and with almost 4M trainable parameters consisting of a linear connection and a Soft Prompt, we can adapt a frozen 7B model and reach an accuracy of 54.5\% on VQA v2, with unconstrained open-ended generation. We validate the effectiveness of the approach with Images, Video, and Audio modalities. 
This direct finetuning setup has several advantages; (a) training data/compute efficiency, (b) attains generally higher performance than pretrain-zeroshot setup, (c) easy to adapt to new tasks, modalities or other LLMs, where no costly pretraining is needed. However, the mechanism proposed in eP-ALM can be adapted in a straightforward manner to this setup.

Even though the results are still far from the state-of-the-art approaches that train most of the model parameters on much more data, the extremely small percentage of trainable parameters (0.06\%) and the increasing scores with model size and compute make the work promising towards finding an intermediate point, between extremely efficient and extremely inefficient approaches, which is hopefully closer to the former. 

The method has some limitations, which we illustrate in the appendix. In general, the model struggles to capture fine-grained details in the images, favors coherent generation over factual one, might hallucinate some objects not present in the image, and lacks common sense reasoning. Our approach inherits most of the limitations and biases of pretrained models, especially the LM, and training only a few adaptation parameters does not seem to avoid the transfer of these biases. Finally, the model is trained with next token prediction and is able to produce coherent text, however, it is still not clear how this paradigm can lead to real reasoning capabilities.

\section{Acknowledgments} This work was partly supported by ANR grant VISA DEEP (ANR-20-CHIA-0022), and HPC resources of IDRIS under the allocation 2022-[AD011013415] and 2023-[AD011013415R1] made by GENCI. The authors would like to thank Theophane Vallaeys for fruitful discussion.

{\small
\bibliographystyle{ieee_fullname}
\bibliography{epalm}
}

\appendix
\section*{Appendix}

The appendix is organized as follows; in Sec. \ref{sec:ap_implem}, we give more implementation details about the experiments that we conduct. We illustrate and explain the different variants of eP-ALM in Sec. \ref{sec:ap_variants}. We compare eP-ALM to other approaches following the pretrain-zeroshot setup (Sec.\ref{sec:pretrain_zeroshot}). We then present more ablation studies on image-text and video-text tasks in Sec. \ref{sec:ap_ablation}. Finally in Sec. \ref{sec:ap_limit} we show some qualitative results and discuss the limitation of the proposed approach. 
\begin{figure*}[h]
    \centering
    \resizebox{\linewidth}{!}{
    \includegraphics[width=\linewidth]{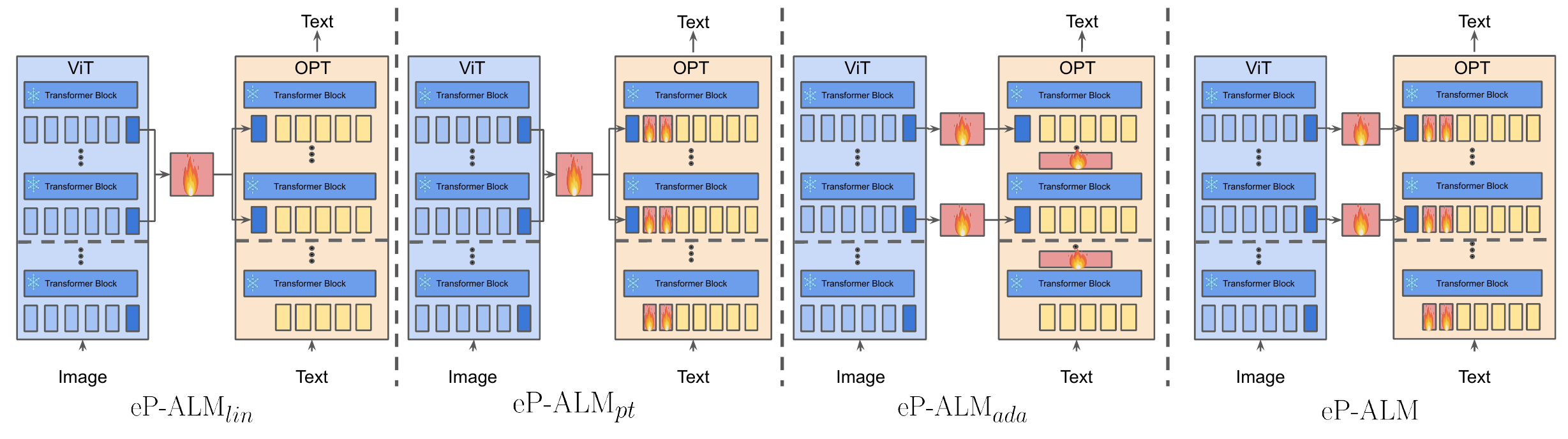}
    }
    \caption{Illustration of the different variants of eP-ALM; eP-ALM$_{lin}$ is the most efficient variant that only trains the linear projection layer, eP-ALM$_{pt}$ adds trainable Soft Prompts (\emph{i.e} Prompt Tuning),  and eP-ALM$_{ada}$ replaces the Soft Prompt in eP-ALM (last figure) with trainable Adapters. All models extract the [CLS] tokens from the last layers of ViT and prepend/replace them in the last layers of OPT.}
    \label{fig:versions}
\end{figure*}

\section{Implementation Details}
\label{sec:ap_implem}
We use OPT-2.7B in our final model. We extract the [CLS] tokens of the last 6 layers of perceptual encoders and prepend them, after linear projection, to the text tokens of the last 12 layers of the OPT. Note that we replace the previous [CLS] with the new one to keep the same number of tokens. We finetune with the classical cross-entropy loss used to train the original OPT for VQA and Captioning. We use the AdamW optimizer with a learning rate of 1e-5 warmed up to 2e-5 then decreased to 1e-6 using a cosine scheduler. For \textbf{Adapters}, we use sequential Adapters after self-attentions and feedforward layers with a downsampling factor of 8 and ReLU activation. For \textbf{Soft Prompt}, we implement it as a linear embedding layer that takes numbers from 0 to the length of the prompt (here 10). We experiment also with adding an MLP after the prompts as done with other approaches \cite{sung2022vl}. We use the prompt with MLP for most of the experiments as we find that it gives slightly better results. The soft prompt and adapters are trained with a fixed lr of 1e-5.
\textbf{eP-ALM$_{pt}$-L} is trained with a light-weight prompt (only trainable tokens without MLP), starting learning rate of 2e-4 and a fixed learning rate of 1e-3 for the prompt with a total batch size of 16.

\paragraph{VQA/GQA:} we use a special token for VQA ($'</a>'$) to separate the question from the answer.  We train for 8 epochs with a batch size of 64 (128 for GQA) and an image resolution of 224. Training our approach with OPT-2.7B for VQA v2 can be done on a single V100 GPU 32GB for 1.8 days (as the perceptual encoder is frozen, saving its output tokens can save a lot of training time). For Few-shot experiments, we train longer (for 64 epochs) with a higher starting learning rate (1e-4 warmed up to 2e-4 and decreased to 1e-5). Those marked by a $^*$ are trained for 100 epochs as in PromptFuse \cite{liang2022modularpromptfuse}.

\paragraph{Image Captioning}  we train for 8 epochs with a batch size of 64 and an image resolution of 224.

\paragraph{Video QA:}  we sample randomly 8 frames of resolution 224x224 for each video and train for 25 epochs with a batch size of 32. For Zero-Shot experiments, we train only for 4 epochs with starting learning rate of 1e-4. We use only the spatial self-attention of TimeSformer to train on VQA v2.

\paragraph{Video Captioning:}  we sample randomly 16 frames of resolution 224x224 for each video and train for 25 epochs with a batch size of 64.

\paragraph{Audio Captioning} we train for 30 epochs with frequency and time masking of 24 and 96 respectively. The mel bins is 128 and the audio length is 1024. Batch size 32. For Deep Prompt, we inject new soft prompts in all 32 blocks of OPT (each with length 10).

\section{eP-ALM Variants}
\label{sec:ap_variants}
We detail the different variants proposed in this paper (here we consider ViT-B/16 and OPT-350M for simplicity). These variants are illustrated in Fig.\ref{fig:versions}:

\paragraph{eP-ALM$_{lin}$:} we extract the [CLS] tokens from the last 6 layers of the frozen ViT and inject them in the last 12 layers of the frozen OPT. To reduce inference cost, for each couple of layers (here 2), we replace the previous [CLS] with the new one (thus only increasing the number of tokens by 1 the whole process). All visual [CLS] tokens are projected by one trainable linear projection layer (shared) to fit their dimension to that of the OPT.

\paragraph{eP-ALM$_{pt}$:} we augment eP-ALM$_{lin}$ with Prompt Tuning, which consists of prepending trainable tokens (\emph{i.e}, soft prompt) to the input of the LM. This might help the model to adapt well to the new task by providing context to the text input. For the sake of efficiency, we prepend only 10 learnable tokens.

\paragraph{eP-ALM:} while one linear projection is appealing, it might not be able to capture all the particularity of different [CLS] tokens. To overcome this, we use different projections for each [CLS], while keeping the soft prompt. 

\paragraph{eP-ALM$_{ada}$:} another alternative to Prompt Tuning are Adapters. We follow other approaches \cite{eichenberg2021magma} and add sequentially one adapter module (downsample, activation then upsample) after self-attention and feedforward layers in all the blocks of OPT. While this might give better results, it adds a significant number of trainable parameters.

\section{Pretrain-Zeroshot Setup}
\label{sec:pretrain_zeroshot}
The focus of this work is on direct finetuning, where we propose an efficient cross-modal interaction mechanism with low data regime. However, the proposed mechanism can be adapted in a straightforward manner to the pretrain-zeroshot evalution setup. In this section, we show the effecitveness of eP-ALM with zero-shot evaluation after pretraining on CC3M. Specifically, we pretrain eP-ALM$_{pt}-L$ on CC3M for 4 epochs (which takes ~35hours on 2 gpus V10032GB), and evalute on COCO \cite{lin2014microsoftcoco} and NoCaps (all) \cite{agrawal2019nocaps} datasets. We experiment with ViT-L, initialized from ImageNet and CLIP. 

Tab. \ref{tab:zeroshot} show a comparison with other approaches. Without using CLIP encoder, eP-ALM significantly outperforms other work on both datasets. Using CLIP (which is trained to produce visual features aligned to text) reduces the improvement gap, where eP-ALM still outperforms all baselines on B@4 and METEOR metrics. This validates that in case of unaligned visual encoder (\emph{e.g.}, pretrained on ImageNet) our cross-modal interaction mechanism is efficient to align both modalities. 

Note that, eP-ALM is significantly more efficient than LimBEr and MAGMA, as they train more parameters for very long time ($\sim >$ 670 GPUhs). As MAGMA is trained on a lot more data, we compare with MAGMA trained on CC3M obtained from \cite{merullo2022linearlylimber}.

\begin{table*}[h]
    \small
    \centering  
 \setlength\tabcolsep{4pt}
    \resizebox{0.9\linewidth}{!}{%
    \begin{tabular} {lcccccccccc}
        \toprule        
       \multirow{2}{*}{Method} & Trainable params. & CLIP Enc.  &  \multicolumn{3}{c}{COCO} & \multicolumn{3}{c}{NoCaps} \\
     &  &   & B@4 &  CIDEr & METEOR  & B@4 &  CIDEr & METEOR \\
  \midrule
  MAGMA (NFRN) \cite{eichenberg2021magma} & 243M & \xmark  & 8.2 & 22.4 & - & 4.5 & 20.9 & - \\
  
  LimBEr (NFRN) \cite{merullo2022linearlylimber} & 8.4M & \xmark  & - & 36.2 & - & - & 28.5 & -  \\ 
    eP-ALM$_{pt}$-L & 4.2M  & \xmark & 11.50 &  42.47 & 15.44 & 12.53 & 36.79 & 15.50     \\ 
  \midrule
  MAGMA (on CC3M from \cite{merullo2022linearlylimber}) & 243M & \cmark & 9.7 & 47.5 & 14.6 & - & 38.7 & - \\
  LimBEr  \cite{merullo2022linearlylimber} & 12.5M & \cmark & 12.6 & 54.9 & 16.1 & - & 43.9 & - \\ 
  FROMAGe \cite{koh2023groundingfromage} & 4.2M & \cmark &  9.65 & - & 11.53 & - & - & - \\

eP-ALM$_{pt}$-L & 4.2M  & \cmark & 12.97 & 51.29  & 16.23 & 13.30 & 39.5 & 15.55    \\ 
 \bottomrule
    \end{tabular}
    }
    \vspace{2ex}
    \caption
    {
    \footnotesize   
Zero-shot comparison on COCO and NoCaps, after pretraining on CC3M (2.84M examples).
        }
    \label{tab:zeroshot}

\end{table*}

\section{Ablation Study}
\label{sec:ap_ablation}
Here we present an additional ablation study.
\subsection{Image-Text}

\paragraph{Training All Parameters}

Here we investigate how much gain we can obtain by unfreezing the pretrained models. We experiment on VQA v2 with eP-ALM. Table \ref{tab:unfreeze} shows that finetuning the pretrained models in our eP-ALM gives slight improvement, despite a large number of trainable parameters. Note that, we find that using a very small learning rate (lr=1e-7) is the only option (while keeping an lr of 1e-5 for the connectors) to unfreeze these models without significant degradation. 

\begin{table}[!t]
    \small
    \centering  
 \setlength\tabcolsep{4pt}
    \begin{tabular} {c c c |  c  }
        \toprule        
     \multicolumn{2}{c}{Trainable Models} & LM &  VQA v2  \\
     VM & LM & size & test Acc. \\
     \midrule
 \xmark & \xmark &350M &   33.08  \\
 \xmark & \cmark &  350M & 35.44  \\
 \cmark & \cmark &  350M & 35.47   \\
    \bottomrule
    \end{tabular}
    \vspace{2ex}
    \caption
    {
    \footnotesize   
        Ablation study: we study how much gain we can obtain by also training the pretrained vision and language models. We see slight improvement by training the pretrained models. 
        }
    \label{tab:unfreeze}

\end{table}

\subsection{Video-Text}

\paragraph{Video Encoder:} here we compare different encoders to process the videos. We compare the TimeSformer \cite{bertasius2021spaceTimeSformer} that has both spatial and temporal attention and trained for video classification with a simple baseline, ViT trained on ImageNet, that ignores the temporal dynamics. For ViT, we take the average of [CLS] tokens of the processed frames while for TimeSformer we consider the single [CLS] token. Table \ref{tab:ab_vid} shows that using video-specific encoders gives significantly better results for video captioning. In addition, we find that using 16 frames instead of 8 gives slight improvement.

\begin{table}[h]
    \small
    \centering  
 \setlength\tabcolsep{4pt}
    \resizebox{0.6\linewidth}{!}{%
    \begin{tabular} {l@{\hspace{24pt}} c  c }
        \toprule        
     \multirow{2}{*}{Method}  &  \multicolumn{2}{c}{MSRVTT} \\ \cmidrule(){2-3}
     & CIDEr & B@4  \\
  \midrule
 ViT-B Avg.    &  17.96  &   12.77        \\
 ViT-B Avg. (16 f)    & 17.82  &   12.85      \\
 \midrule
  TimeSformer & 20.11 & 13.53  \\
  TimeSformer (16 f) & 20.58 & 14.12    \\

    \bottomrule
    \end{tabular}
    }
    \vspace{2ex}
    \caption
    {
    \footnotesize   
        Ablation (Caption) MSRVTT Caption.
        }
    \label{tab:ab_vid}

\end{table}

\paragraph{Injection and Extraction level of [CLS] tokens:} here we show the importance of leveraging the hierarchical representation in both  the video encoder and language model. Table \ref{tab:ab_cls_vid} shows the results on MSVD-QA. We show that keeping the interaction between cross-modal tokens to the last layers (layers 19 to 31) of the OPT leads to significantly better results. Extracting several tokens from different tokens of the TimeSformer gives slight improvement. However, using hierarchical video transformers \cite{li2021improvedmvitv2, liu2022videoswin} might lead to better results. We noticed also that Adapters generally give better results than Prompt Tuning, this might be because when training on videos we sample randomly some frames, which prevents the model to overfit in the case of small datasets.

\begin{table}[!h]
    \small
    \centering  
 \setlength\tabcolsep{4pt}
    \resizebox{0.9\linewidth}{!}{%
    \begin{tabular} {lc c @{\hspace{25pt}}  c  }
        \toprule        
     Adaptation & \multicolumn{2}{c}{[CLS] tokens} &  MSVD-QA  \\
     approach & from encoder layers & to OPT layers &  test Acc. \\
     \midrule
 \multirow{4}{*}{Soft Prompt} & 12  & 1  & 13.49    \\
 & 12   & 1 to 31 & 27.16   \\
 & 12 & 19 to 31 & 30.86   \\
 & 6 to 12 & 19 to 31 & 31.18    \\
 \midrule
 \multirow{3}{*}{Adapters} & 12  & 1  & 12.40    \\
 & 12  & 1 to 31 & 34.86    \\
 & 12   & 19 to 31 & 35.94   \\

    \bottomrule
    \end{tabular}
    }
    \vspace{2ex}
    \caption
    {
    \footnotesize   
        Ablation study: we investigate the extraction and injection position of [CLS] tokens for Video QA. 
        }
    \label{tab:ab_cls_vid}

\end{table}

\subsection{Audio-Text}

\paragraph{Comparison with different variants.}
Here we compare different variants of our approach to different baselines for audio captioning. We evaluate on AudioCaps dataset \cite{audiocaps}, the largest benchmark for Audio Captioning. We train with mel spectrograms of 128 bins and frequency and time masking with a batch size of 8.  

To the best of our knowledge, no prior work has been proposed to adapt LM for audio-text tasks. However, there is some recent work adapting audio models using parameter-efficient techniques, such as Deep Prompts and Adapters \cite{kim2022integrated}. Tab. \ref{tab:audiocap_ablation} shows a comparison with different approaches. We find that prepending the audio tokens to the input of OPT does not give reasonable performance. To investigate this more, we train another baseline where the audio tokens are concatenated in the last 12 layers of OPT (eP-ALM$_{l19-31+Adapter}$ and eP-ALM$_{l19-31+DeepPT}$). This leads to significant improvement. 

\begin{table}[h]
    \small
    \centering  
 \setlength\tabcolsep{4pt}
    \resizebox{0.7\linewidth}{!}{%
    \begin{tabular} {lc@{\hspace{25pt}}c@{\hspace{10pt}}c}
        \toprule        
     \multirow{2}{*}{Method} & Trained & \multicolumn{2}{c@{\hspace{10pt}}}{AudioCaps} \\ \cmidrule(lr{10pt}){3-4}
     & param (\%) & CIDEr & B@4  \\
  \midrule
    B$_{Adapter}$ & 3.76 \%  & 2.96  & - \\
    \midrule
    
    eP-ALM$_{l19-31+Adapter}$ & 3.76 \% & 31.17  & 8.09      \\
    eP-ALM$_{l19-31+DeepPT}$  & 0.93 \%  & 32.57  &  10.66     \\

    eP-ALM$_{pt}$ & 0.55 \% & 35.17 & 10.73  \\
    eP-ALM & 0.90 \% & \textbf{37.14} & \textbf{11.37}   \\

    \bottomrule
    \end{tabular}
    }
    \vspace{2ex}
    \caption
    {
    \footnotesize   
        Comparison with other work for Audio Captioning on AudioCaps Test set.
        }
    \label{tab:audiocap_ablation}

\end{table}

\paragraph{Time and Frequency Masking:} following other approaches \cite{gong21b_interspeech_ast, park2019specaugment} we train eP-ALM with time and frequency masking on AudioCaps. Table \ref{tab:comp_2_7b_aud} shows that masking significantly helps, however, using too much masking hurt the performance. 
\begin{table}[h]
    \small
    \centering  
 \setlength\tabcolsep{4pt}
    \resizebox{0.7\linewidth}{!}{%
    \begin{tabular} {c c @{\hspace{25pt}} c c }
        \toprule        
     \multicolumn{2}{c@{\hspace{25pt}}}{Masking Window}  & \multicolumn{2}{c}{AudioCaps} \\ \cmidrule(){1-2}  \cmidrule(){3-4}
    Time & Frequency & CIDEr & B@4  \\
  \midrule
 256 &  64  & 33.94  &  10.21     \\
 192 &  48  & 35.67  &  10.40     \\
 96 &  24  & 37.14  &  11.37     \\
 0 &  0  &  36.01 &   10.23    \\
    \bottomrule
    \end{tabular}
    }
    \vspace{2ex}
    \caption
    {
    \footnotesize   
        Ablation Study: time and frequency masking help for Audio Captioning.
        }
    \label{tab:comp_2_7b_aud}

\end{table}

\section{Limitations}
\label{sec:ap_limit}
\begin{figure*}[h]
    \centering
    \resizebox{0.9\linewidth}{!}{
    \includegraphics[width=\linewidth]{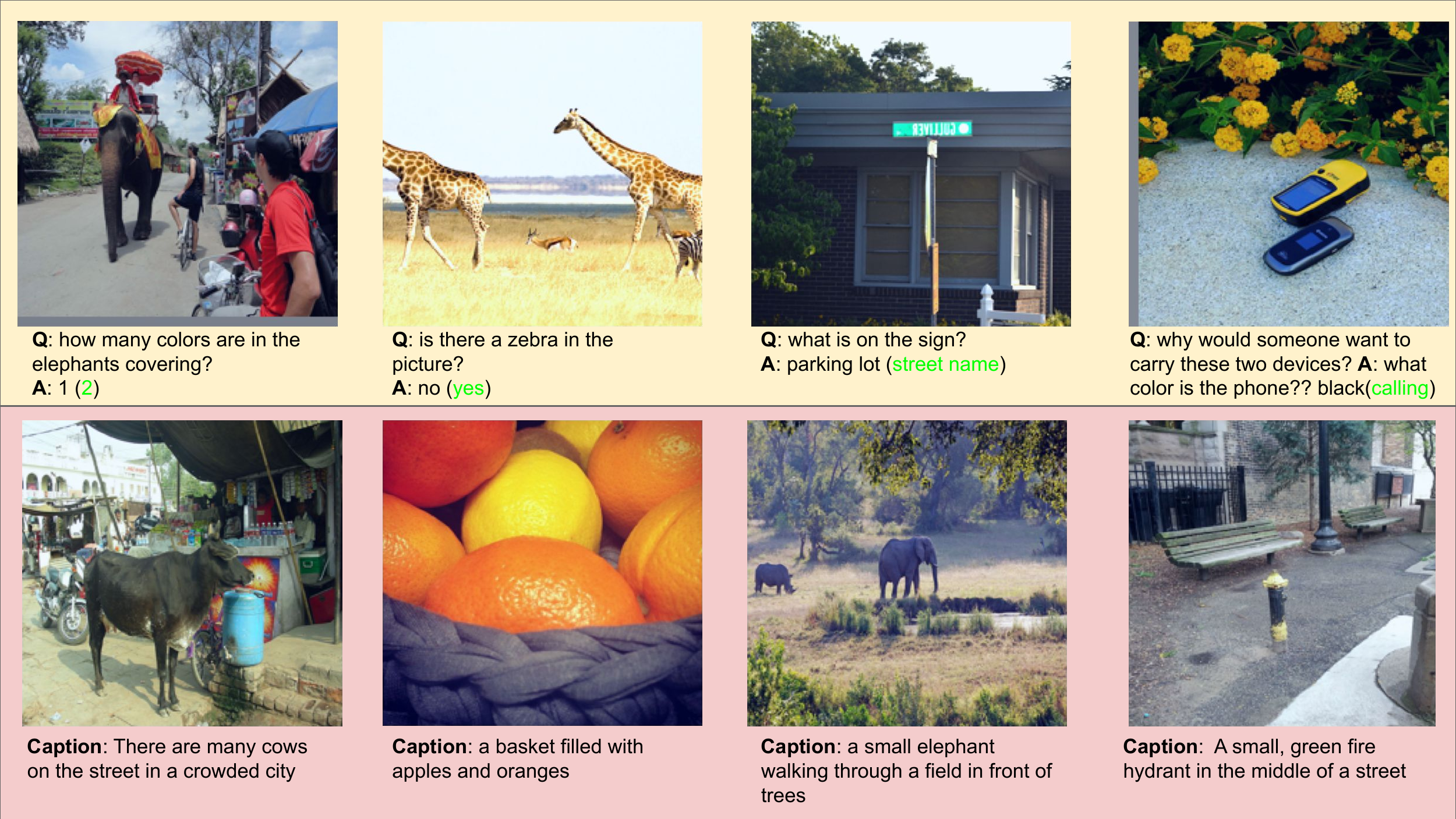}
    }
    \caption{Illustration of some limitations of eP-ALM: the model struggles to capture fine-grained details, favors coherence over factual responses, hallucinates some objects, and lacks common sense reasoning. Ground truth answers are highlighted in green. Results obtained using multinomial sampling.}
    \label{fig:qual_limit}
\end{figure*}

Even though we show appealing results for very efficient training, the method has several limitations, which we illustrate in Fig. \ref{fig:qual_limit}. For VQA, we can notice that the model is unable to capture fine-grained details in the images (\emph{e.g.}, number of colors and the zebra in the first 2 examples), which might be due to constraining the interaction with the vision model through the [CLS] tokens, that generally capture global information about the image. In the case of hard questions, the model favors a coherent generation of a relevant question  followed by its correct answer, instead of answering the main question ("A: what
color is the phone?? black" in example 3).

For Captioning, the model seems to favor outputting a coherent sentence, even though it is not entirely correct ("many" cows in a "crowded" city). Secondly, the model might hallucinate some objects that do not appear in the image ("apples" in example 2). Finally, the model lacks common sense reasoning, making him unable to understand that elephants are not small, and being far from the camera does not change this fact (example 3).

Our approach inherits most of the limitations and biases of pretrained models, especially the LM, and training only a few adaptation parameters does not seem to avoid the transfer of these biases. Finally, the model is trained with next token prediction and is able to produce coherent text, however, it is still not clear how this paradigm can lead to real reasoning capabilities.

\begin{figure*}[h]
    \centering
    \resizebox{0.9\linewidth}{!}{
    \includegraphics[width=\linewidth]{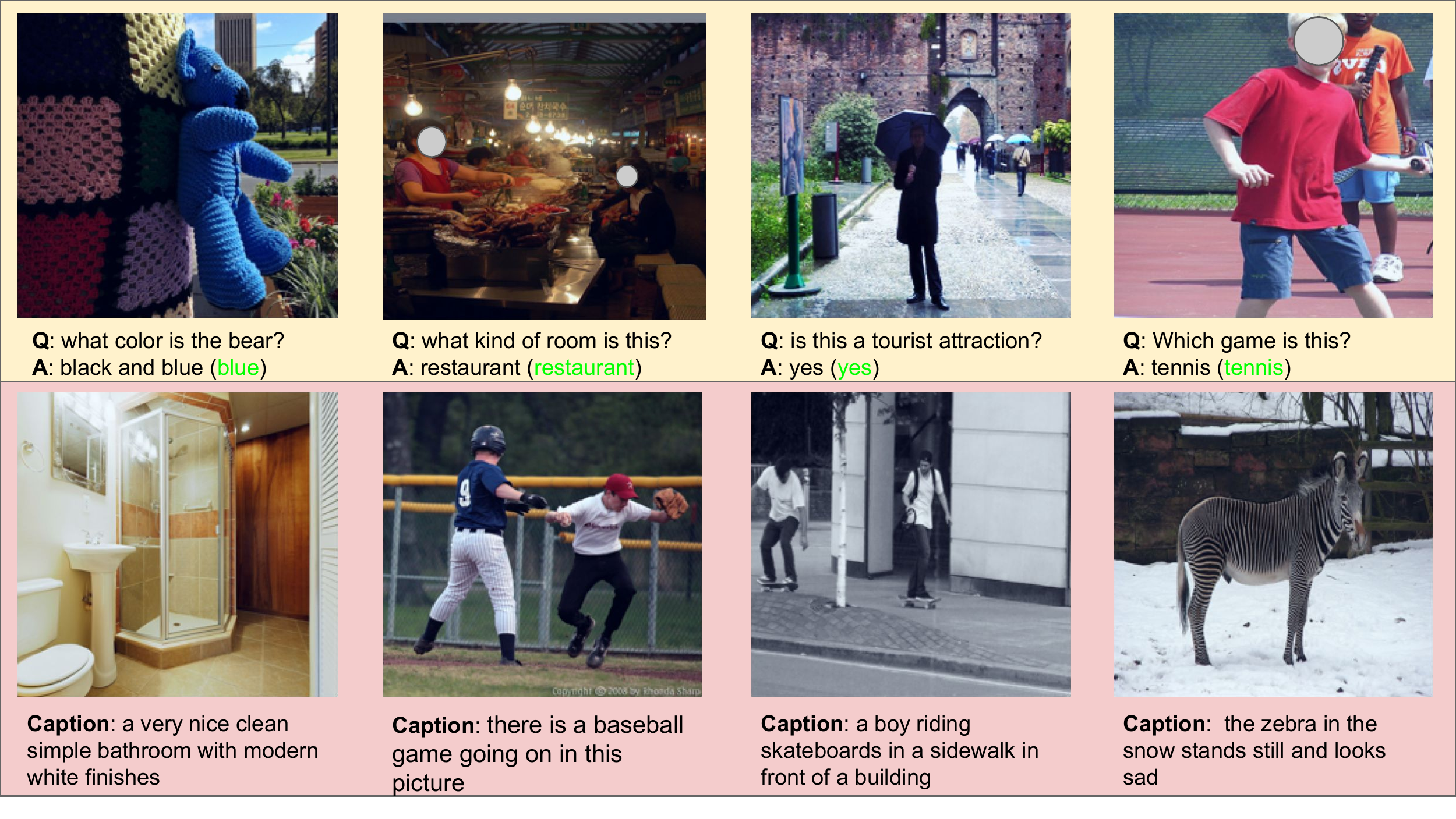}
    }
    \caption{Qualitative results of eP-ALM: the model is able to generate accurate answers and coherent descriptions of the image. Ground truth answers are highlighted in green. Results obtained using multinomial sampling.}
    \label{fig:qual}
\end{figure*}

\end{document}